\documentclass[journal]{Style/IEEEtran}
\usepackage{microtype}
\usepackage{float}
\usepackage{graphicx}
\usepackage[caption=false,font=footnotesize]{subfig}
\usepackage{booktabs} 
\usepackage{bbding}
\usepackage{multirow}
\usepackage{bm}
\usepackage[table,xcdraw]{xcolor}
\usepackage[normalem]{ulem}
\useunder{\uline}{\ul}{}

\usepackage{algorithm}
\usepackage{algorithmic}
\usepackage{amsmath}

\usepackage{graphics}
\usepackage{epsfig}

\usepackage{amsthm}

\newtheorem{definition}{Definition}

\allowdisplaybreaks[4]

\usepackage{amssymb}
\usepackage{mathtools}

\usepackage{capt-of}
\usepackage{afterpage}

\definecolor{forward}{RGB}{88,182,229}
\definecolor{backward}{RGB}{241,136,112}
\definecolor{model}{rgb}{1.0, 0.75, 0.0}
\definecolor{azure}{rgb}{0, 0.5, 1.0}

\usepackage{hyperref}

\def\ddefloop#1{\ifx\ddefloop#1\else\ddef{#1}\expandafter\ddefloop\fi}

\def\ddef#1{\expandafter\def\csname v#1\endcsname{\ensuremath{\boldsymbol{#1}}}}
\ddefloop ABCDEFGHIJKLMNOPQRSTUVWXYZabcdefghijklmnopqrstuvwxyz\ddefloop

\def\ddef#1{\expandafter\def\csname v#1\endcsname{\ensuremath{\boldsymbol{\csname #1\endcsname}}}}
\ddefloop {alpha}{beta}{gamma}{delta}{epsilon}{varepsilon}{zeta}{eta}{theta}{vartheta}{iota}{kappa}{lambda}{mu}{nu}{xi}{pi}{varpi}{rho}{varrho}{sigma}{varsigma}{tau}{upsilon}{phi}{varphi}{chi}{psi}{omega}{Gamma}{Delta}{Theta}{Lambda}{Xi}{Pi}{Sigma}{varSigma}{Upsilon}{Phi}{Psi}{Omega}{ell}\ddefloop

\def\ddef#1{\expandafter\def\csname bb#1\endcsname{\ensuremath{\mathbb{#1}}}}
\ddefloop ABCDEFGHIJKLMNOPQRSTUVWXYZ\ddefloop

\renewcommand{\paragraph}[1]{\noindent\textbf{#1.}}

\begin{document}

\title{Dealing with Structure Constraints in \\ Evolutionary Pareto Set Learning}

\author{
Xi Lin,
        Xiaoyuan Zhang,
        Zhiyuan Yang,
        Qingfu Zhang,~\IEEEmembership{Fellow,~IEEE}
\thanks{Xi Lin, Xiaoyuan Zhang, Zhiyuan Yang, and Qingfu Zhang are with the Department of Computer Science, City University of
Hong Kong, Hong Kong SAR, China. Email: xi.lin@my.cityu.edu.hk, xzhang2523-c@my.cityu.edu.hk, zhiyuyang4-c@my.cityu.edu.hk, qingfu.zhang@cityu.edu.hk (Corresponding Author: Qingfu Zhang).}
}

\maketitle

\begin{abstract}

In the past few decades, many multiobjective evolutionary optimization algorithms (MOEAs) have been proposed to find a finite set of approximate Pareto solutions for a given problem in a single run, each with its own structure. However, in many real-world applications, it could be desirable to have structure constraints on the entire optimal solution set, which define the patterns shared among all solutions. The current population-based MOEAs cannot properly handle such requirements. In this work, we make the first attempt to incorporate the structure constraints into the whole solution set by a single Pareto set model, which can be efficiently learned by a simple evolutionary stochastic optimization method. With our proposed method, the decision-makers can flexibly trade off the Pareto optimality with preferred structures among all solutions, which is not supported by previous MOEAs. A set of experiments on benchmark test suites and real-world application problems fully demonstrates the efficiency of our proposed method.     

\end{abstract}


%
\IEEEpeerreviewmaketitle

\section{Introduction}
\label{sec_intro}

%
%
%
%

\IEEEPARstart{M}{ultiobjective} optimization problems (MOPs) naturally arise in many real-world applications, from engineering design~\cite{tanabe2020easy} to machine learning~\cite{lin2019pareto,lu2020neural}. In this paper, we consider the following continuous multiobjective optimization problem:
\begin{align}
\min_{\vx \in \mathcal{X}} \vF(\vx) = (f_1(\vx),f_2(\vx),\cdots,
f_m(\vx)),
\label{eq_mop}
\end{align}
where $\vx$ is the decision variable in the decision space $\mathcal{X} \subset \bbR^n$ and $\vf: \bbR^n \rightarrow \bbR^m$ consists of $m$ objective functions. Very often, these objective functions conflict one another, and no single solution $\vx$ can optimize all the objectives at the same time. Instead, there is a set of Pareto optimal solutions with different optimal trade-offs among objectives. Under mild conditions, the set of all Pareto optimal solutions (called the Pareto set) and its image in the objective space (called the Pareto front) are both ($m-1$)-dimensional piecewise continuous manifolds in the decision space and objective space, respectively~\cite{hillermeier2001generalized, zhang2008rm}. In this work, we assume that all objective functions  are black-box without explicit gradient information.

Over the past three decades, many multiobjective optimization algorithms have been proposed to find a single or a finite set of Pareto optimal solutions with different trade-offs among all objectives~\cite{ehrgott2005multicriteria, miettinen2012nonlinear}. Among them, the multiobjective evolutionary algorithms (MOEA)~\cite{zitzler1999evolutionary,deb2001multi,zhou2011multiobjective} is a widely-used methodology to produce a finite set of approximate Pareto optimal solutions in a single run. However, a finite set of solutions cannot always approximate the whole Pareto set well. In many real-world applications, it could be tough for decision-makers to extract their desired information from a finite set of approximate Pareto solutions to make final decisions. Many model-based methods have been proposed to assist or improve traditional MOEAs, such as surrogate-assisted optimization~\cite{knowles2006parego,zhang2010expensive,chugh2016surrogate}, model-based preference assignment~\cite{wu2018learning,liu2019adapting,liu2020adaptive}, hypervolume approximation~\cite{shang2022hv,shang2022hvc} and Pareto set/front approximation from a finite set of solutions~\cite{rakowska1991tracing,hillermeier2001generalized,zhang2008rm,hartikainen2011constructing,giagkiozis2014pareto,cheng2015multiobjective}.

Recently, a one-stage Pareto set learning (PSL) framework has been proposed to approximate the whole Pareto set by a single model~\cite{lin2022pareto_moco,lin2022pareto_expensive,guo2023approximation, zhang2024hypervolume}, which does not require approximate Pareto optimal solutions in advance. With the gradient-based training method, it can learn the Pareto optimal set for deep multi-task learning~\cite{lin2020controllable,navon2021learning,ruchte2021scalable}, neural multiobjective combinatorial optimization~\cite{lin2022pareto_moco, wang2023multiobjective, chen2024neural}, expensive multiobjective optimization~\cite{lin2022pareto_expensive, lu2024you}, and multiobjective molecular design~\cite{jain2023multi,zhu2024sample, liu2023multi}. Different approaches have been proposed to improve the efficiency and performance of Pareto set learning~\cite{chen2022multi, lin2024smooth, ye2024evolutionary, ye2024data}. However, these methods are not designed for solving black-box optimization problems, and none of them takes the Pareto set structure into consideration for the set learning process.

In many real-world applications, it could be desirable to add some structure constraints on the optimal solution set. For example, in multiobjective modular design~\cite{rai2003modular,sinha2018pareto}, the production cost can be significantly reduced if the solutions (i.e., design schemes) for different scenarios share some common components. Innovization proposed and advocated by Deb et al.\cite{deb2003unveiling,deb2006innovization,deb2014integrated,bandaru2015generalized, guha2023regemo} aims at mining useful patterns among different Pareto optimal solutions for decision making, which is conducted as \textit{post-optimization analysis}. In this paper, we propose a different method to actively incorporate decision-maker-preferred structure patterns into the whole solution set via evolutionary Pareto set learning (EPSL). In a sense, our method can be regarded as a model-based approach for innovization that actively incorporates predefined structure patterns \textit{as prior}. Using our proposed method, the Pareto optimality of the solution set can now be traded off with the desirable structure constraint. We believe it is necessary to go beyond Pareto optimality in multiobjective optimization and consider other user requirements on common solution patterns in the optimization process.

Our main contributions can be summarized as follows:

\begin{itemize}

\item We make a first attempt to add different decision-maker-defined structure constraints on the whole solution set for multiobjective optimization.

\item We propose a general set optimization formulation and a lightweight evolutionary algorithm to learn the Pareto set with different structure constraints.

\item We experimentally compare our proposed method with other representative MOEAs on different multiobjective engineering optimization problems to validate its ability to deal with structure constraints. The results confirm its effectiveness and usefulness of EPSL for solving MOPs.

\end{itemize}

\section{Preliminaries: Multi-Objective Optimization and Evolutionary Pareto Set Learning}
\label{sec_psl}

\begin{figure*}[t]
    \centering
    \includegraphics[width= 1 \linewidth]{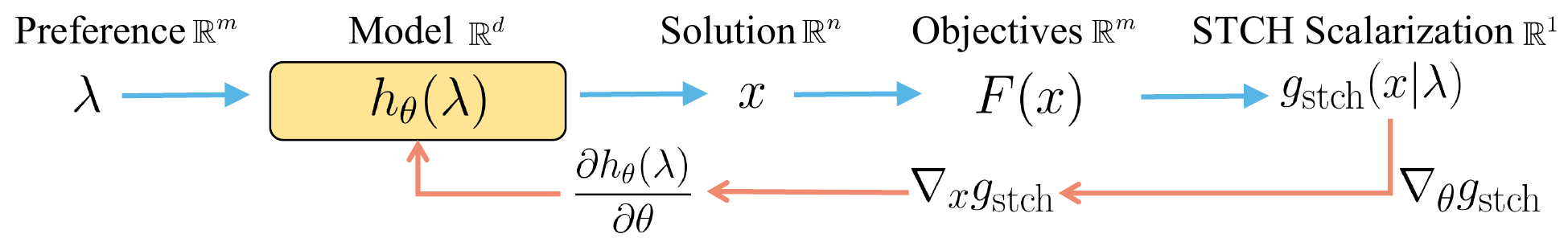}
    \caption{\textbf{Evolutionary Pareto Set Learning (EPSL):} \textbf{(a)} \textbf{The forward pass ($\color{forward} \rightarrow$)} starts from the preference $\vlambda \in \bbR^m$ to the model $h_{\vtheta}(\vlambda)$ with parameter $\theta \in \bbR^d$ to the solution $\vx \in \bbR^n$ to the objectives $\vF(\vx) \in \bbR^m$ and finally to the preference-conditioned subproblem scalar $g_{\text{stch}}(\vx|\vlambda) \in \bbR^1$. The learnable \textcolor{model}{\textbf{model parameter}} $\theta$ is the only variable to optimize in this model. \textbf{(b)} \textbf{The backward gradient $\nabla_{\vtheta} g_{\text{stch}}$ ($\color{backward} \leftarrow$)} from the subproblem scalar $g_{\text{stch}}(\vx|\vlambda) \in \bbR^1$ with respect to the learnable model parameter $\vtheta \in \bbR^d$ can be decomposed into a low-dimensional gradient $\nabla_{\vx} g_{\text{stch}}$ and a high-dimensional Jacobian matrix $\frac{\partial h_{\vtheta}(\vlambda) }{\partial \vtheta}$, where the first term can be efficiently estimated by evolution strategies and the second term can be directly calculated by backpropagation.}
    \label{fig_model_EPSL}
\end{figure*}

In this section, we first review some basic concepts and the Pareto set learning model recently proposed by us~\cite{lin2022pareto_moco,lin2022pareto_expensive}, and then introduce a lightweight evolutionary model training method for black-box multiobjective optimization problem.

\subsection{Basic Definitions}

For an multiobjective optimization problem (\ref{eq_mop}), we have the following definitions:

\begin{definition}[Pareto Dominance]
Let $\vx^{a},\vx^{b} \in \mathcal{X}$ be two candidate solutions for the multiobjective optimization problem~(\ref{eq_mop}), $\vx^{a}$ is said to dominate $\vx^b$, denoted by $\vx^{a} \prec \vx^{b}$ if and only if $f_i(\vx^{a}) \leq f_i(\vx^{b}), \forall i \in \{1,...,m\}$ and $f_j(\vx^{a}) < f_j(\vx^{b}), \exists j \in \{1,...,m\}$.
\end{definition}

\begin{definition}[Strict Dominance]
Let $\vx^{a},\vx^{b} \in \mathcal{X}$, $\vx^{a}$ is said to strictly dominate $\vx^{b}$, denoted by $\vx^{a} \prec_{\text{strict}} \vx^{b}$, if and only if $f_i(\vx^{a}) < f_i(\vx^b), \forall i \in \{1,...,m\}$.
\end{definition}

\begin{definition}[Pareto Optimality] A solution
$\vx^{\ast} \in \mathcal{X}$ is Pareto optimal if there does not exist $\vx \in \mathcal{X}$ such that $ \vx \prec \vx^{\ast}$. The set of all the Pareto optimal solutions is called the Pareto set, denoted by PS, and its image $\vf(PS)$ in the objective space is called the Pareto front (PF).
\end{definition}

\begin{definition}[Weakly Pareto Optimality]
A solution $\vx^{\prime} \in \mathcal{X}$ is weakly Pareto optimal if there does not exist $\hat \vx \in \mathcal{X}$ such that $\hat \vx \prec_{\text{strict}} \vx^{\prime}$. The weakly Pareto set/front can be defined accordingly.
\end{definition}

For a continuous MOP (\ref{eq_mop}), under mild conditions, the Pareto set is an $(m-1)$-dimensional manifold which consists of an infinite number of solutions~\cite{hillermeier2001generalized, zhang2008rm, miettinen2012nonlinear}. In other words, it is natural and feasible to use a math model to represent the PS.

\subsection{Aggregation Methods}

Let $\vlambda \in \vDelta$ be a weight vector on the simplex $\vDelta = \{\vlambda=(\lambda_1, \ldots, \lambda_m)^T \in \bbR^m_{+}| \sum_{i=1}^{m} \lambda_j = 1 \}$, aggregation methods~\cite{miettinen2012nonlinear} combine all the individual objectives into one single objective function and optimize it to obtain one single Pareto optimal solutions. The most widely-used weighted-sum aggregation works as follows:
\begin{eqnarray}
g_{\text{ws}}(\vx|\vlambda) =  \sum_{i=1}^{m} \lambda_i f_i(\vx).
\label{ws_decomposition}
\end{eqnarray}
It is well-known that the weighted-sum aggregation method is not able to find some Pareto optimal solutions when the PF is not convex. Many other aggregation methods have been developed to overcome this shortcoming over the past several decades~\cite{das1998normal,zhang2007moea,liu2013decomposition, wang2016localized,jiang2017scalarizing}. For example, the Tchebycheff (TCH) aggregation function is defined as:
\begin{eqnarray}
 g_{\text{tch}}(\vx|\vlambda) =  \max_{1 \leq j \leq m} \{ \lambda_j(f_j(\vx) - (z^*_j - \varepsilon))\},
\label{tch_decomposition}
\end{eqnarray}
where $z_{j}^* =\min_{\vx \in \mathcal{X}} f_j(\vx)$ is an ideal value for the $i$-th objective, $\varepsilon > 0$ is a small positive component, and $(z_j^* - \varepsilon)$ is also called an (unattainable) utopia value for the $i$-th objective. The necessary and sufficient condition for Tchebycheff aggregation to obtain the whole Pareto set is:

\textbf{Theorem 1 (\cite{choo1983proper}).} \textit{A feasible solution $x \in \mathcal{X}$ is weakly Pareto optimal for the multiobjective optimization problem if and only if there is a weight vector $\vlambda > 0$ such that $\vx$ is an optimal solution of the Tchebycheff aggregation subproblem~(\ref{tch_decomposition}).}
In other words, all weakly Pareto solutions $\vx^*$ can be found by solving the Tchebycheff scalarized subproblem with a specific weight vector $\vlambda$. This result suggests that we can use the weight vector $\vlambda \in \vDelta$ to parameterize the PS.

In this paper, we use a recently developed smooth Tchebycheff (STCH) aggregation approach~\cite{lin2024smooth} in our proposed method:
\begin{equation}
g_{\text{stch}}(\vx|\vlambda, \mu) = \mu \log \left(\sum_{j=1}^m e^{\frac{\lambda_j(f_j(\vx) - (z^*_j - \varepsilon))}{\mu}} \right),
\label{stch_decomposition}
\end{equation}
where $z^*_j$ and $\varepsilon$ are the ideal value and a small positive component as defined in the Tchebycheff aggregation, and $\mu$ is the smoothing parameter. In this work, we always set $\mu = 0.1$ and denote $g_{\text{stch}}(\vx|\vlambda) = g_{\text{stch}}(\vx|\vlambda, \mu = 0.1)$ for brevity. According to \cite{lin2024smooth}, $g_{\text{stch}}(\vx|\vlambda, \mu)$ is a smooth approximation to the classical Tchebycheff aggregation. It enjoys a faster convergence rate for gradient-based optimization methods, while having many good theoretical properties for multiobjective optimization. In this work, we show that STCH aggregation also has a better overall performance for evolutionary stochastic optimization. Our proposed method can be readily generalized to other aggregation approaches.

Many classic optimization methods have been proposed to find a single Pareto optimal solution by optimizing a single-objective aggregation function~\cite{ehrgott2005multicriteria,miettinen2012nonlinear}, while multiobjective evolutionary algorithms (MOEAs)~\cite{zitzler1999evolutionary,deb2001multi,zhou2011multiobjective} can find a number of different Pareto optimal solutions in a single run. Among them, the decomposition-based MOEAs (MOEA/D)~\cite{zhang2007moea,li2009multiobjective} use an aggregation method to decompose a MOP into a finite number of single-objective subproblems with different weight vectors and solve them in a collaborative manner. This paper generalizes MOEA/D to produce optimal solutions for all (might be infinite) weight vectors via a single model.

\subsection{Pareto Set Model}

In Pareto set learning, the Pareto set can be represented as: 
\begin{eqnarray}
\vx = h_{\vtheta}(\vlambda),
\label{model}
\end{eqnarray}
where $\vlambda \in \vDelta$, and $h_{\vtheta}: \vDelta \rightarrow \mathcal{X}  $ is a model with learnable parameter $\vtheta \in \Theta$. Our task is to learn an optimal $\vtheta^*$ such that  $\vx^* = h_{\vtheta^*}(\vlambda)$ is the optimal solution to
\begin{eqnarray}
\min_{\vx \in  \mathcal{X} } g_{\text{stch}}(\vx |\vlambda), \hfill \quad \forall \vlambda \in \vDelta.
\label{set_model_tch}
\end{eqnarray}
Let $P(\vlambda)$ be a user-specific probability distribution for $\vlambda$, then our task can be defined as:
\begin{eqnarray}
\min_{\vtheta \in \vTheta} L(\vtheta)=E_{\vlambda \sim P(\cdot)} g_{\text{stch}}(h_{\vtheta}(\vlambda)|\vlambda).
\label{eq_expectation_problem}
\end{eqnarray}
In this paper, $h_{\vtheta}$ is set to be a two-layer fully-connected neural network where $\vtheta$ is its model parameters, and well-developed neural network training algorithms can be naturally used for learning the Pareto set model. This set model formulation naturally generalizes the decomposition-based MOEA (MOEA/D)~\cite{zhang2007moea} from a finite population to a parameterized math model. With the learned model, decision-makers can readily obtain approximate Pareto optimal solutions for any preferences.

We would like to make the following remarks on the Pareto set learning problem (\ref{eq_expectation_problem}):
\begin{itemize}
    \item If $P(\vlambda)$ is a uniform distribution on a finite number of weight vectors, problem (\ref{eq_expectation_problem}) is equivalent to the task of MOEA/D, that is, to optimize a finite number of aggregated functions at the same time.

    \item Problem (\ref{eq_expectation_problem}) is for optimizing model parameter $\vtheta$ instead of decision variable $\vx$. Even when objectives $f_i$ are not differentiable or their gradients are not available, it is still possible to compute or estimate the gradient of $L(\vtheta)$. Therefore, efficient gradient search methods can be used for optimizing problem (\ref{eq_expectation_problem}).

    \item $P(\vlambda)$ can be set to suit the decision-maker's preferences. For example, if the decision-maker is interested in only a preference subset, $P(\vlambda)$ can be set in such a way that its supporting set is this preference subset in problem (\ref{eq_expectation_problem}).

    \item Although this work focuses on continuous MOPs, our modeling approach can also tackle combinatorial problems by reformulating them into single-objective continuous model learning~\cite{lin2022pareto_moco}.

\end{itemize}

\subsection{Evolutionary Stochastic Optimization}
\label{subsec_esgd}

It is very difficult to directly solve the optimization problem (\ref{eq_expectation_problem}) since the expected term in $L(\vtheta)$ cannot be exactly computed in most cases. Treating it as a stochastic optimization problem~\cite{bottou1998online, nemirovski2009robust}, we can use stochastic gradient descent methods to learn the Pareto set model~\cite{lin2020controllable, lin2022pareto_moco,lin2022pareto_expensive}. At each iteration, $L(\vtheta)$ is approximated via Monte Carlo sampling:
\begin{equation}
\tilde{L}(\vtheta) = \frac{1}{N} \sum_{i=1}^{N} g_{\text{stch}}(h_{\vtheta}(\vlambda^i)|\vlambda^i),
\label{eq_mc_sample}
\end{equation}
where $\{\vlambda^1, \vlambda^2, \cdots, \vlambda^N\}$ are $N$ independent identically distributed (i.i.d) samples from $P(\cdot)$. Clearly, $\tilde{L}(\vtheta)$ is an unbiased estimation of $L(\vtheta)$. When $g_{\text{stch}}(h_{\vtheta}(\vlambda)|\vlambda)$ is well-behaved~\cite{nemirovski2009robust} for all $\vlambda \sim P(\cdot)$, we can have:
\begin{align}
 \nabla_{\vtheta} L(\vtheta) & = \nabla_{\vtheta} E_{\vlambda \sim P(\cdot)} [g_{\text{stch}}(h_{\vtheta}(\vlambda)|\vlambda)] \\
 &= E_{\vlambda \sim P(\cdot)} \nabla_{\vtheta}[g_{\text{stch}}(h_{\vtheta}(\vlambda)|\vlambda)]
\end{align}
where the expectation and differentiation are swapped. Then
\begin{equation}
\nabla_{\vtheta} \tilde{L}(\vtheta)=\frac{1}{N} \sum_{i=1}^{N} \nabla_{\vtheta} [g_{\text{stch}}(h_{\vtheta}(\vlambda^i)|\vlambda^i)]
\end{equation}
can be used for approximating $\nabla_{\vtheta} L(\vtheta)$.

For each fixed $\vlambda$, $\nabla_{\vtheta} [g_{\text{stch}}(h_{\vtheta}(\vlambda)|\vlambda)]$ can be computed using the chain rule:
\begin{equation}
\nabla_{\vtheta} g_{\text{stch}}(h_{\vtheta}(\vlambda)|\vlambda) = \frac{\partial h_{\vtheta}(\vlambda) }{\partial \vtheta} \cdot \nabla_{\vx} g_{\text{stch}}(\vx|\vlambda)|_{\vx= h_{\vtheta}(\vlambda)}.
\end{equation}
When $h_{\vtheta}(\vlambda)$ is a neural network and $\vtheta$ is its parameters as in this paper, the Jacobian matrix $\frac{\partial h_{\vtheta}(\vlambda) }{\partial \vtheta}$ can be easily computed via the backpropagation algorithm. A brief illustration of the forward and backward pass of the model can be found in Fig.~\ref{fig_model_EPSL}.

Like other MOEAs, we assume that the multiobjective optimization problem is black-box, and no analytic formula exists for computing $\nabla_{\vx} g_{\text{stch}}(\vx|\vlambda)$. This work uses a simple evolution strategy (ES)~\cite{hansen2001completely, beyer2002evolution,li2020evolution} to estimate the gradient with respect to the solution $\vx$. A similar approach has been discussed in \cite{lin2023continuation} for single-objective continuation path learning. We first sample $K$ independent $n$-dimensional random vector $\{\vu_1, \ldots, \vu_K\}$, and then approximate $\nabla_{\vx} g_{\text{stch}}(\vx|\vlambda)$ by
\begin{align}
 &\overline{\nabla}_{\vx} g_{\text{stch}}(\vx|\vlambda)= \frac{1}{\sigma K} \sum_{k=1}^{K} ( g_{\text{stch}}(\vx+ \sigma \vu_k|\vlambda) - g_{\text{stch}}(\vx|\vlambda)) \cdot \vu_k,
 \label{eq_simple_es}
\end{align}
where $\sigma>0$ is a control parameter. This gradient approximation method is also closely related to the Gaussian smoothing approach~\cite{nesterov2017random,gao2022generalizing}. Following \cite{gao2022generalizing}, this work independently samples each $\vu_k \sim ((B_{0.5} - 0.5) / 0.5)$ where $B_{0.5}$ follows the Bernoulli distribution with probability $0.5$. More advanced sampling and estimation methods can also be adopted into our framework.

\begin{algorithm}[t]
	\caption{Evolutionary Stochastic Optimization for EPSL}
	\label{alg_epsl_framework}
 	\begin{algorithmic}[1]
    	\STATE \textbf{Input:} Model $\vx = h_{\vtheta^0}(\vlambda)$ with Initialized Parameter $\vtheta^0$, Number of MC Samples $N$, Number of Gaussian Samples $K$
		\FOR{$t = 1$ to $T$}
		   \STATE Sample $N$ weight vectors $\{\vlambda^1, \cdots, \vlambda^N\}$ from $P(\vlambda)$
		   \STATE Calculate the estimated gradient for each $\vlambda^i$ with $K$ samples as in (\ref{eq_simple_es_shaping}) and (\ref{eq_chain_rule_estimation})
		   \STATE Update $\vtheta^t$ with evolutionary gradient descent as in (\ref{eq_sgd_es})
		\ENDFOR	
		\STATE \textbf{Output:} Model $\vx = h_{\vtheta^T}(\vlambda)$
 	\end{algorithmic}
\end{algorithm}

In real-world applications, the objective functions $f_i(\vx)$ could have quite different scales, which will lead to inconsistent magnitudes of the value $g_{\text{stch}}(\vx|\vlambda)$ for different preferences $\lambda$. To address this issue, we apply the rank-based fitness shaping approach~\cite{hansen2001completely, wierstra2014natural} to further stabilize the optimization process with the following gradient approximation:
\begin{align}
 &\overline{\nabla}_{\vx} g_{\text{stch}}(\vx|\vlambda)= \frac{1}{\sigma K} \sum_{k^{\prime}=1}^{K} r_k^{\prime} \cdot \vu_{k^{\prime}}.
 \label{eq_simple_es_shaping}
\end{align}
To obtain the fitness $r_k^{\prime}$, we first sort the sampled vectors $\{\vu_k\}_{k=1}^K$ by their values $\{g_{\text{stch}}(\vx+ \sigma \vu_k|\vlambda)\}_{k=1}^K$, and then assign a predefined set of normalized values $r_1 \leq \cdots \leq r_K$ to them according to their ranks. In this work, we let the normalized values be a set of evenly spaced values on $[-0.5, 0.5]$. For example, for five sampled vectors $(\vu_1, \vu_2, \vu_3, \vu_4, \vu_5)$ sorted by their values $g_{\text{stch}}(\vx+ \sigma \vu_k^{\prime}|\vlambda)\}$ in increasing order, their normalized values should be $(-0.5, -0.25, 0, 0.25, 0.5)$.   

With the ES-based gradient estimation, we can approximate the gradient for $\tilde{L}(\vtheta)$ by:
\begin{equation}
\overline{\nabla}_{\vtheta} \tilde{L}(\vtheta) =\frac{1}{N}\sum_{i=1}^{N} \frac{\partial h_{\vtheta}(\vlambda^i) }{\partial \vtheta} \cdot \overline{\nabla}_{\vx} g_{\text{stch}}(\vx|\vlambda^i)|_{x=h_{\vtheta}(\vlambda)}.
\label{eq_chain_rule_estimation}
\end{equation}
Let $\vtheta^t$ be the model parameter at iteration $t$, we propose to update it with the simple gradient descent:
\begin{equation}
\vtheta^{t+1} = \vtheta^{t}- \eta^t \overline{\nabla}_{\vtheta} \tilde{L}(\vtheta^t),
\label{eq_sgd_es}
\end{equation}
where $\eta^t > 0$ is the step size. This can be regarded as a combination of the evolutionary gradient search (EGS)~\cite{salomon1998evolutionary,arnold2007evolutionary} and the stochastic optimization methods~\cite{robbins1951stochastic, bottou1998online, nemirovski2009robust}. We call this method as Evolutionary Pareto Set learning (EPSL) algorithm and summarize it in \textbf{Algorithm~\ref{alg_epsl_framework}}.

\section{Structure Constraints for Solution Set}
\label{sec_epsl_structure}

\subsection{Structure Constraints}

To incorporate structure constraints on optimal solution sets in multiobjective optimization, we define a general set optimization problem for MOP (\ref{eq_mop}). Let $\vS \subset \mathcal{X}$ be the set we want to find and $Q(\vS)$ be a quality metric of $\vS$ with regards to the objective vector $\vF(\vx)$, our goal is to
\begin{equation}
\begin{array}{l l}
\mbox{min} & Q(\vS),\\
\mbox{s.t.} &  \mbox{Some constraints on $\vS$}.
\end{array}
\label{GSP}
\end{equation}
We call this problem the multiobjective set optimization problem (MSOP). Many existing optimization tasks for dealing with the MOP (\ref{eq_mop}) can be regarded as special cases of the MSOP. Two typical examples are:
\begin{itemize}
    \item When the constraint on $\vS$ is that any two different elements in $\vS$ cannot dominate each other, and $Q$ is the hypervolume of $\vS$,  the MSOP (\ref{GSP}) becomes the problem of finding all the Pareto optimal solutions.
    \item When the constraint on $\vS$ is that $\vS$ can be represent as a Pareto set model (\ref{model}) and $Q$ is defined as the expected values over all valid subproblems (\ref{eq_expectation_problem}), the MSOP (\ref{GSP}) is the Pareto set learning problem studied in the above section.
\end{itemize}

It should be pointed out that some effort has been made to view multiobjective optimization problems as set optimization problems. Particularly, performance indicator based algorithms~\cite{zitzler2004indicator,beume2007sms,bader2011hype} aim to optimize a performance metric of a solution set with finite solutions. However, they do not consider structure constraints. We consider several typical structure constraints on the whole solution set and hence also on the Pareto set model $h_{\vtheta}(\vlambda)$ in the following subsections.

\subsection{Shared Component Constraint}
\begin{figure}[t]
    \centering
    \includegraphics[width= 0.85 \linewidth]{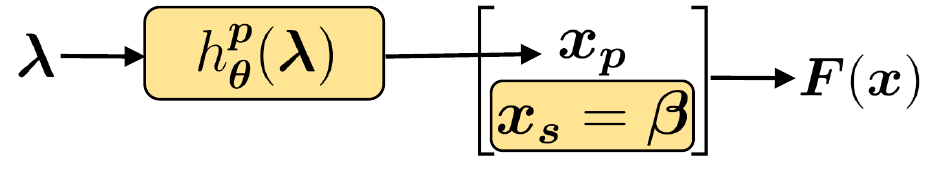}
    \caption{EPSL Model for solution set with shared components. The variables to optimize are the model parameters $\vtheta$ and part of the decision variables $\vx_{\vs}=\vbeta$ that shared by all trade-off solutions.}
    \label{fig_model_constant}
\end{figure}

To support multiobjective modular design, decision-makers may require that all the solutions in $\vS$ share some common components. More precisely, let $\vs \subset \{1, \ldots, n \}$, $\vp=\{1, \ldots, n \} \backslash \vs$, $\vx_{\vs}$ and $\vx_{\vp}$ be the subvector of $\vx$ associated with the index sets $\vs$ and $\vp$, respectively. The shared component constraint on $\vS$ with respect to $\vs$ is that the $\vx_{\vs}$-values of all the solutions in $\vS$ are the same. If we consider this constraint in Paerto set learning (\ref{model}), then the model can be expressed as
\begin{equation}
\vx_{\vs}=\vbeta, \;\;\; \vx_{\vp}=h^{\vp}_{\vtheta}(\vlambda)
\label{SCC1}
\end{equation}
where $\vbeta$ and $\vtheta$ are two sets of parameters to learn, $h^{\vp}$ is a map from $\vDelta$ to $R^{|\vp|}$. The model structure is illustrated in Fig.~\ref{fig_model_constant}.

For the model with shared component constraint (\ref{SCC1}), the optimization problem (\ref{eq_expectation_problem}) becomes:
\begin{eqnarray}
\min_{\vtheta, \vbeta} L(\vtheta, \vbeta)=\bbE_{\vlambda \sim P(\vlambda)} g_{\text{stch}}( \vx_{\vs} = \vbeta, \vx_{\vp} = h^{\vp}_{\vtheta}(\vlambda)|\vlambda).
\label{eq_expectation_problem_shared_variables}
\end{eqnarray}
The gradient of $L(\vtheta, \vbeta)$ can be computed as follows:
\begin{align}
 &\nabla_{\vtheta} L(\vtheta,\vbeta)=\bbE_{\vlambda \sim P(\vlambda)} \nabla_{\vtheta}g_{\text{stch}}( \vx_{\vs} = \vbeta, \vx_{\vp} = h^{\vp}_{\vtheta}(\vlambda)|\vlambda),\\
 &\nabla_{\vbeta} L(\vtheta,\vbeta)=\bbE_{\vlambda \sim P(\vlambda)} \nabla_{\vbeta}g_{\text{stch}}( \vx_{\vs} = \vbeta, \vx_{\vp} = h^{\vp}_{\vtheta}(\vlambda)|\vlambda)
\end{align}
where
\begin{align}
&\nabla_{\vtheta} g_{\text{stch}} = \frac{\partial  h^{\vp}_{\vtheta}(\vlambda) }{\partial \vtheta} \cdot \nabla_{\vx_{\vp}} g_{\text{stch}}([\vx_{\vs} = \vbeta, \vx_{\vp} =  h_{\vtheta}(\vlambda)]|\vlambda), \\
&\nabla_{\vbeta} g_{\text{stch}} = \nabla_{\vx_{\vs}} g_{\text{stch}}(\vx)|_{\vx_{\vs}=\vbeta}.
\end{align}
It is straightforward to use the proposed evolutionary stochastic optimization method in \textbf{Algorithm~\ref{alg_epsl_framework}} to optimize $\vtheta$ and $\vbeta$ at the same time.

\subsection{Learnable Variable Relationship Constraint}

\begin{figure}[h]
    \centering
    \includegraphics[width= 0.85 \linewidth]{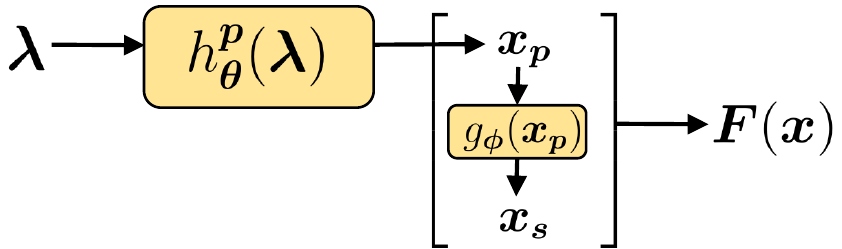}
    \caption{EPSL Model for solution set with learnable variable relationship constraints. The variables to optimize are the model parameters $\vtheta$ and the learnable relation expression $g_{\vphi}$ from some base decision variables $\vx_{\vp}$ to the dependent decision variables $\vx_{\vs}$.}
    \label{fig_model_relation_constraint}
\end{figure}
In some real-world applications such as engineering design problems~\cite{deb2006innovization},  decision-makers may require that the decision variables satisfy some specified relationship. Let $\vx_{\vs}$ and $\vx_{\vp}$ as defined in the previous subsection, we consider the following relationship between $\vx_{\vs}$ and $\vx_{\vp}$:
\begin{equation}
    \vx_{\vs}=g_{\vphi}(\vx_{\vp}),
    \label{relationshipconstraint}
\end{equation}
where $g$ is a function with a learnable parameter $\vphi$, and it characterizes the relationship between $\vx_{\vs}$ and $\vx_{\vp}$. For example, $g$ can be a linear function as studied in \cite{deb2003unveiling}.

When the learnable relation structure (\ref{relationshipconstraint}) is considered, the Pareto set model (\ref{model}) can be expressed as
\begin{equation}
\vx_{\vs}=g_{\vphi}(\vx_{\vp}),\;\;\; \vx_{\vp}=h^{\vp}_{\vtheta}(\vlambda)
\label{SCC2}
\end{equation}
where $\vphi$ and $\vtheta$ are two sets of parameters to learn. The model structure is illustrated in Fig.~\ref{fig_model_relation_constraint}.

For the model with variable relation constraint (\ref{SCC2}), the optimization problem (\ref{eq_expectation_problem}) becomes:
\begin{align}
\min_{\vtheta, \vphi} L(\vtheta, \vphi) = \bbE_{\vlambda \sim P(\vlambda)} g_{\text{stch}}( \vx_{\vs}=g_{\vphi}(\vx_{\vp}), \vx_{\vp}=h^{\vp}_{\vtheta}(\vlambda)|\vlambda).
\label{eq_expectation_problem2constraint}
\end{align}
The gradient of $L(\vtheta, \vphi)$ can be computed as follows:
\begin{align}
&\nabla_{\vtheta} L(\vtheta, \vphi)=\bbE_{\vlambda \sim P(\vlambda)} \nabla_{\vtheta}g_{\text{stch}}( g_{\vphi}(h^{\vp}_{\vtheta}(\vlambda)), h^{\vp}_{\vtheta}(\vlambda)|\vlambda), \\
&\nabla_{\vphi} L(\vtheta, \vphi)=\bbE_{\vlambda \sim P(\vlambda)} \nabla_{\vphi}g_{\text{stch}}( g_{\vphi}(h^{\vp}_{\vtheta}(\vlambda)), h^{\vp}_{\vtheta}(\vlambda)|\vlambda)
\end{align}
where
\begin{align}
\nabla_{\vtheta} g_{\text{stch}} &=\frac{\partial h^p_{\vtheta}(\vlambda) }{\partial \vtheta} \cdot [\nabla_{\vx_{\vp}}g_{\text{stch}}(\vx|\vlambda) \nonumber\\
&+\frac{\partial g_{\phi}(\vx_{\vp}) }{\partial \vx_{\vp}}
\nabla_{\vx_{\vs}} g_{\text{stch}}(\vx|\vlambda)]|_{\vx_{\vs}=g^{\vp}_{\vphi}(\vx_{\vp}), \vx_{\vp}=h^{\vp}_{\vtheta}(\vlambda)}, \\
\nabla_{\vphi} g_{\text{stch}} &= \frac{\partial  g_{\vphi}(\vlambda) }{\partial \vphi} \cdot
\nabla_{\vx_{\vs}} g_{\text{stch}}(\vx|\vlambda)|_{\vx_{\vs}=g^{\vp}_{\vphi}(\vx_{\vp}), \vx_{\vp}=h^{\vp}_{\vtheta}(\vlambda)}.
\end{align}
Our proposed evolutionary stochastic optimization method in \textbf{Algorithm~\ref{alg_epsl_framework}} can be directly used to optimize $\vtheta$ and $\vphi$.

\subsection{Shape Structure Constraint}

\begin{figure}[h]
    \centering
    \includegraphics[width= 0.95 \linewidth]{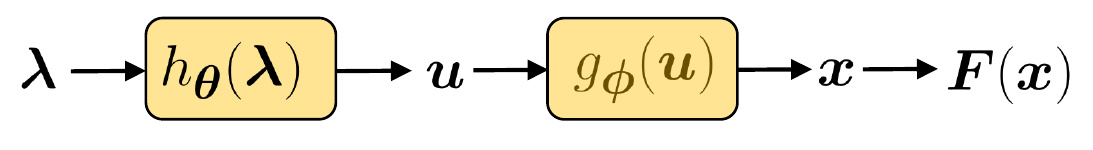}
    \caption{EPSL Model for solution set with shape structure constraint. The variables to optimize are the model parameters $\vtheta$ and the parameters $\vphi$ for the shape function $g_{\vphi}(\vu)$.}
    \label{fig_model_shape_constraint}
\end{figure}

In some multiobjective problems, the decision-makers may require that the solution set has an intrinsic low-dimensional structure or a pre-defined shape. For this purpose, we introduce the compositional set model with shape function $g_{\vphi}$:
\begin{eqnarray}
\vu(\vlambda) = h_{\vtheta}(\vlambda), \quad \vx = g_{\vphi}(\vu(\vlambda)),
\label{set_model_structure}
\end{eqnarray}
where $\vu \in \bbR^d$ is a latent embedding with dimension $d \leq m - 1$, and $\vtheta, \vphi$ are the model parameters to optimize. Fig.~\ref{fig_model_shape_constraint} illustrates the set model architecture with such shape structure constraints. In this model, the essential dimensionality of the solution set can be the same as that of $\vmu$, and its geometrical shape is determined by the function $g_{\vphi}$.

\begin{figure*}[t]
\centering
\subfloat[]{\includegraphics[width = 0.25\linewidth]{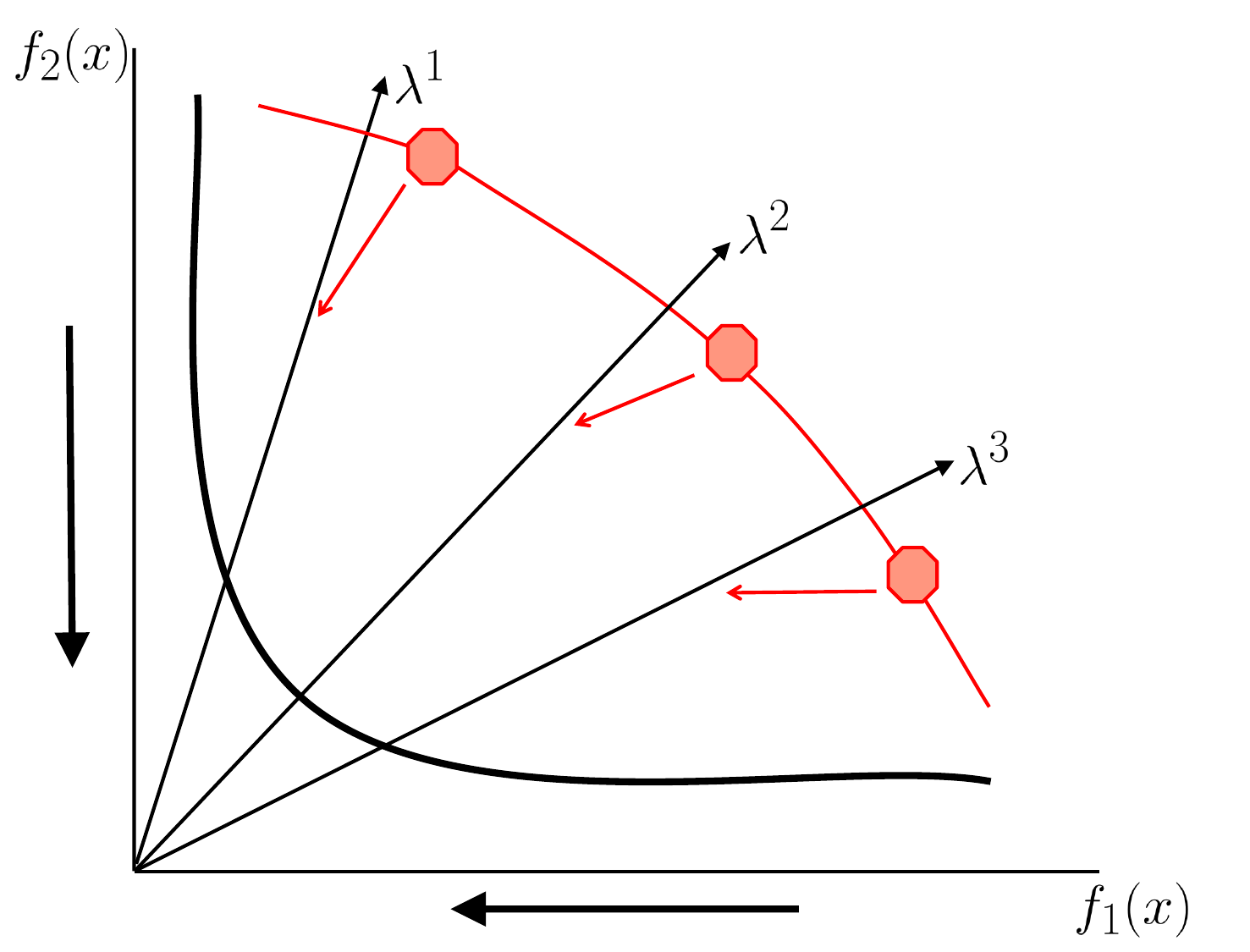}}
\hfill
\subfloat[]{\includegraphics[width = 0.25\linewidth]{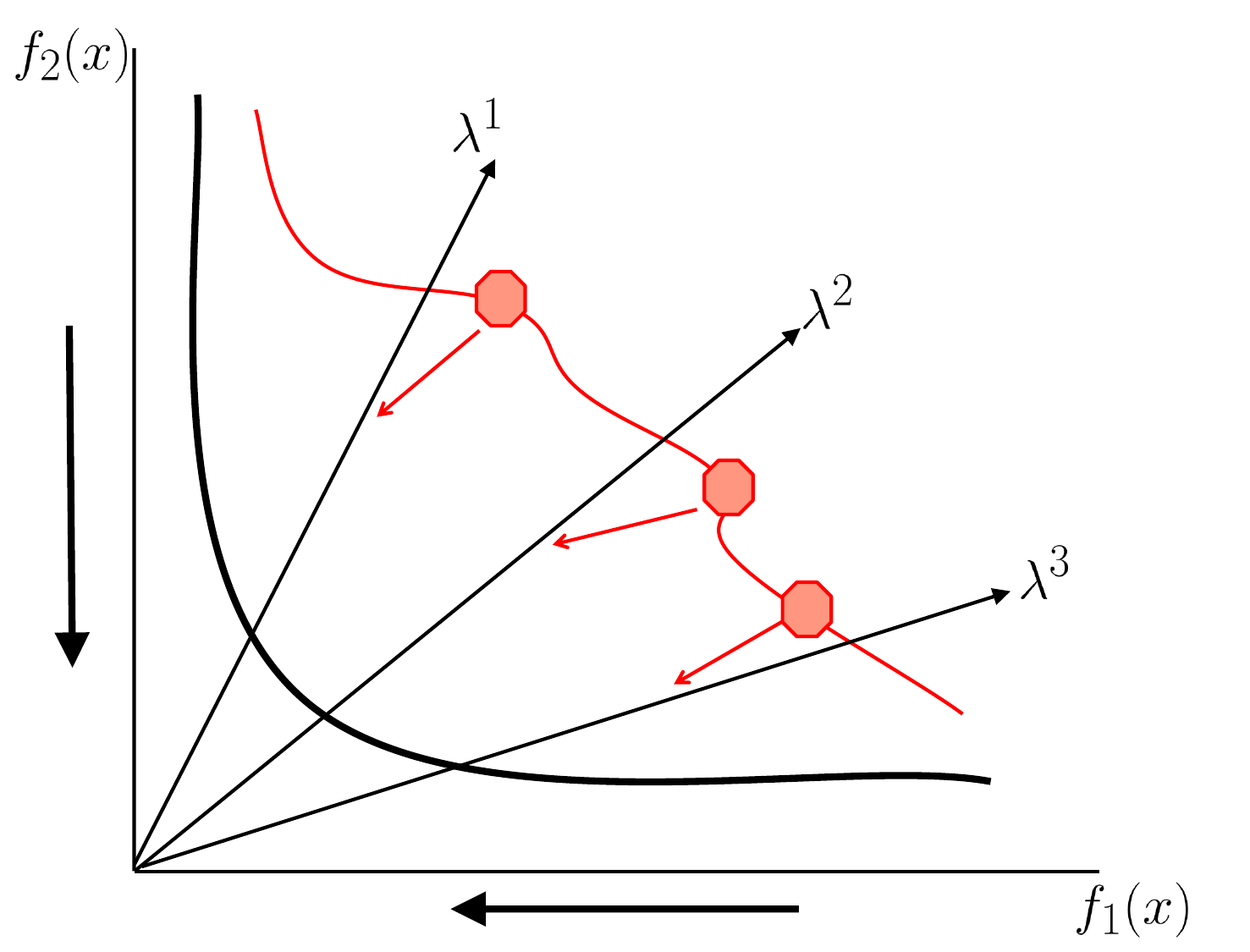}}
\hfill
\subfloat[]{\includegraphics[width = 0.25\linewidth]{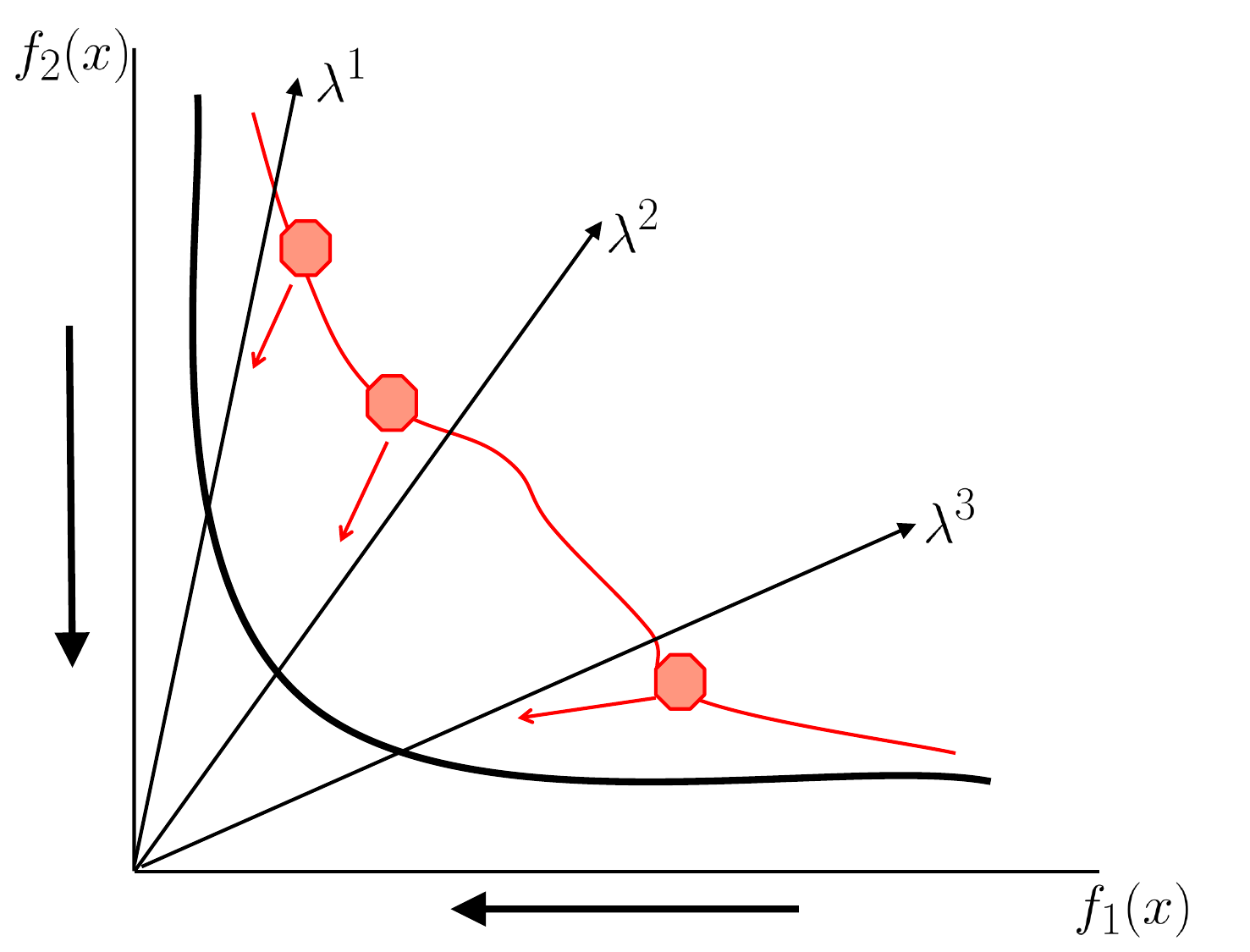}}
\hfill
\subfloat[]{\includegraphics[width = 0.25\linewidth]{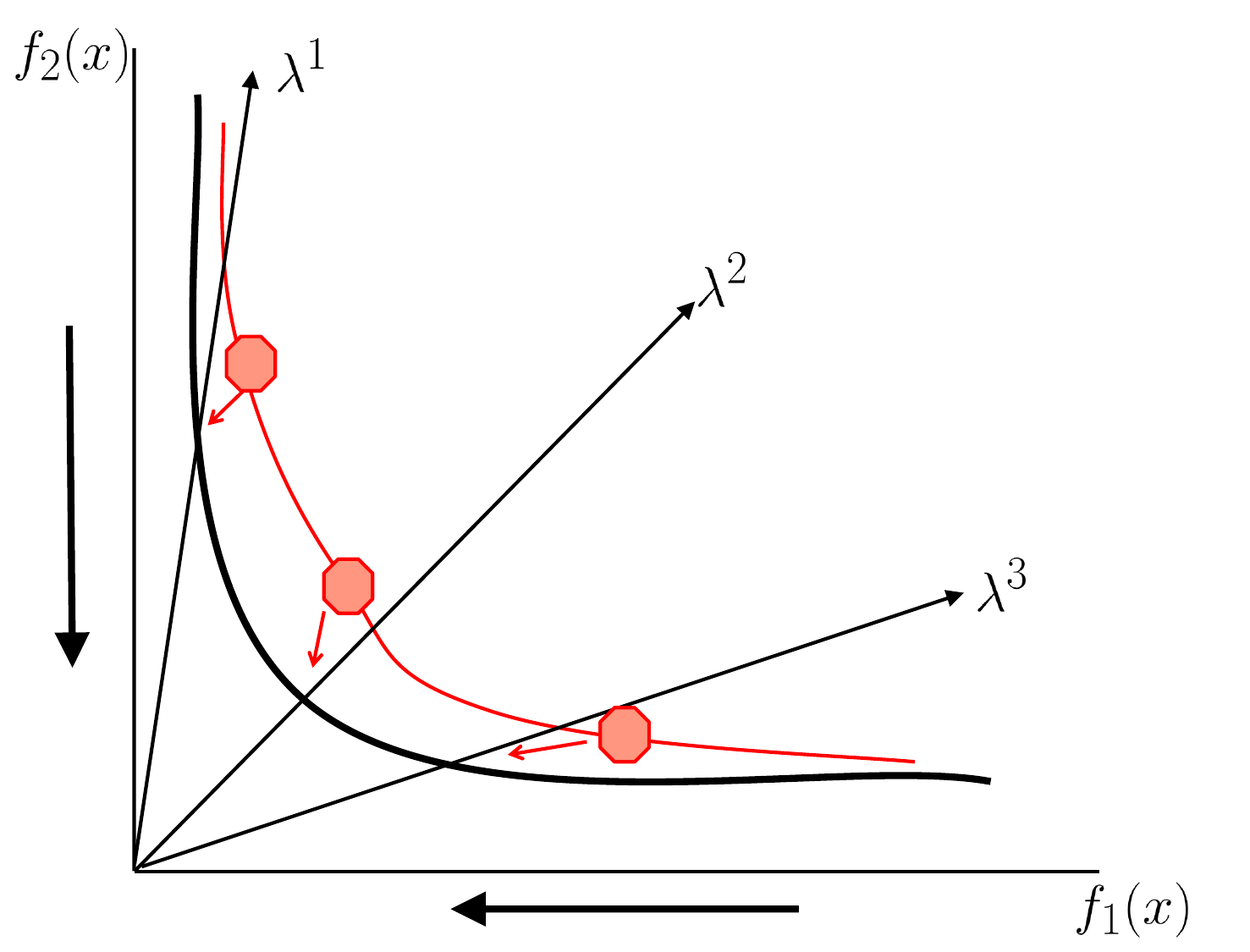}}
\caption{\textbf{Convergence of our Proposed Evolutionary Pareto Set Learning (EPSL) Method:} EPSL gradually pushes the set model to ground truth Pareto set via iteratively reducing the corresponding subproblems values of the sampled solutions. It does \textbf{not} require any Pareto-optimal solution just like the traditional MOEA counterparts which gradually push a set of solutions (e.g., population) to the ground truth Pareto set.}
\label{fig_opt_process}
\end{figure*}

With the shape structure constraints, we have the following optimization problem for set model learning:
\begin{align}
\min_{\vtheta, \vphi} L(\vtheta, \vphi) = &\bbE_{\vlambda \sim P(\vlambda)}
 g_{\text{stch}}(\vx = g_{\vphi}(h_{\vtheta}(\vlambda))|\vlambda).
\label{eq_expectation_constraint_shape}
\end{align}

The gradients of $L(\vtheta, \vphi)$ with respect to the parameters $\vtheta$ and $\vphi$ are:
\begin{align}
&\nabla_{\vtheta} L(\vtheta, \vphi )= \bbE_{\vlambda \sim P(\vlambda)} \nabla_{\vtheta}g_{\text{stch}}( \vx = g_{\vphi}(h_{\vtheta}(\vlambda))|\vlambda), \\
&\nabla_{\vphi} L(\vtheta, \vphi) = \bbE_{\vlambda \sim P(\vlambda)} \nabla_{\vphi}g_{\text{stch}}( \vx = g_{\vphi}(h_{\vtheta}(\vlambda))|\vlambda)
\end{align}
where
\begin{align}
\nabla_{\vtheta} g_{\text{stch}} &=\frac{\partial h_{\vtheta}(\vlambda) }{\partial \vtheta} \cdot [\frac{\partial g_{\phi}(\vu) }{\partial \vu}
\nabla_{\vx} g_{\text{stch}}(\vx|\vlambda)]|_{\vx = g_{\vphi}(h_{\vtheta}(\vlambda))}, \\
\nabla_{\vphi} g_{\text{stch}} &= \frac{\partial  g_{\vphi}(\vu) }{\partial \vphi} \cdot
\nabla_{\vx} g_{\text{stch}}(\vx|\vlambda)|_{\vx_{\vx = g_{\vphi}(h_{\vtheta}(\vlambda))}}.
\end{align}
The parameters $\vtheta$ and $\vphi$ can be optimized with \textbf{Algorithm~\ref{alg_epsl_framework}}.

\begin{figure}[t]
    \centering
    \includegraphics[width= 0.95 \linewidth]{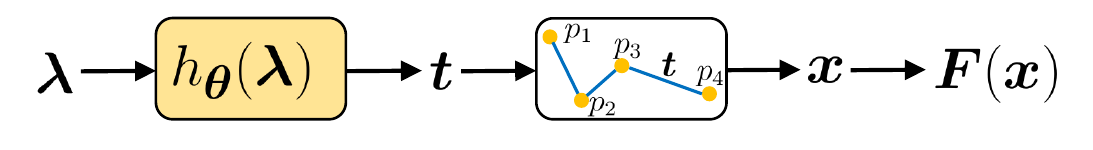}
    \caption{EPSL Model for the solution set with learnable polygonal chain. The variables to optimize are the model parameters $\vtheta$ for the tracer $t_{\vtheta}(\vlambda)$, and the locations for a sequence of vertices $[\vp_1, \vp_2,\ldots,\vp_K]$.}
    \label{fig_model_polygonal_chain}
\end{figure}

The form of shape function $g_{\vphi}(\vx)$ can be provided by decision-makers to model their requirement.  One possible application is to use a simple and explicit expression to approximate the solution set for supporting decision making.
In this paper, we consider a simple polygonal chain structure for bi-objective optimization problem, and leave the investigation on other shape structures to future work.

The polygonal chain (or piecewise linear curve) is a simple way to approximate a space curve for capturing its skeleton~\cite{Bellman1961On, Dunham1986Optimum, Rosin1997Techniques}. A polygonal chain $\vP_c$ with $K-1$ line segments can be defined by a sequence of knee points $[\vp_1, \vp_2, ..., \vp_K]$ called its vertices and a location parameter $t \in [0, K]$ called the tracer. A solution on $\vP_c$ with tracer $t$ can be represented by:
\begin{align}
\vx = \vp_k + (t - k)(\vp_{k+1} - \vp_{k}), \text{ if }  t \in [k,k+1]
\label{eq_polygonal_chain}
\end{align}
for $k = 1, 2, \ldots, K$.
It is straightforward to incorporate the polygonal chain structure into our proposed set model as illustrated in Fig.~\ref{fig_model_polygonal_chain}. Now we have:
\begin{align}
t &= h_{\vtheta}(\vlambda)  \nonumber \\
\vx &= g_{\vphi}(t) = \vp_k + (t(\vlambda) - k)(\vp_{k+1} - \vp_{k}), \nonumber \\
&\text{ if } t(\vlambda) \in [k,k+1]
\label{eq_polygonal_chain_model}
\end{align}
for  $k = 1, 2, \ldots, K$.
The tracer $t = h_{\vtheta}(\vlambda)$ is the learnable latent embedding of the solution set, and the shape function $g_{\vphi}(\cdot)$ is fully parameterized by the learnable vertices $[\vp_1, \vp_2, ..., \vp_K]$. Once the approximate solution set is obtained, it can be fully reconstructed with the $K$ vertices $[\vp_1, \vp_2, ..., \vp_K]$ where the number $K$ is usually small in practice. Decision-makers can readily find any preferred solutions by assigning their preference $\vlambda$.

\subsection{Discussion}

We could like to make the following remarks on the proposed evolutionary Pareto set learning (EPSL) approach with structure constraints.

\begin{itemize}

\item \textbf{No Pareto Optimal Solution is Required:} Our proposed method learns the set model by gradually minimizing the corresponding objective values for different trade-offs, and does \textbf{not} require any Pareto-optimal solutions in advance. For a multiobjective optimization with respect to solution $\vx$, we reformulate it as the set learning problem approach~(\ref{eq_expectation_problem}) potentially with structure constraints~(\ref{eq_expectation_problem_shared_variables}, \ref{eq_expectation_problem2constraint}, \ref{eq_expectation_constraint_shape}), which is an optimization problem with respect to the model parameter $\vtheta$. At each iteration, we randomly sample $N$ different preferences $\{\lambda_i\}_{i=1}^{N}$ and use evolutionary stochastic gradient descent to update the model parameter $\theta$. This procedure directly reduces the smooth Tchebycheff aggregated subproblem values $\sum_{i=1}^{N} g_{\text{stch}}(\vx = h_{\vtheta}(\vlambda^i)|\vlambda^i)$ with respect to the sampled current corresponding solutions $\vx^i = h_{\vtheta}(\vlambda^i)$ predicted by the model as shown in Figure~\ref{fig_opt_process}, and it does \textbf{not} require to explicitly minimize the prediction error between the predicted solutions and the corresponding (unknown) Pareto-optimal solutions. In this way, we can gradually push our Pareto set model to the ground truth Pareto set without any Pareto-optimal solution. 

\item \textbf{Sample Complexity:} Since our EPSL method does not require any Pareto-optimal solution, the training complexity mainly comes from the function evaluations during the optimization process. If the training objective is convex and differentiable, the complexity of simple first-order gradient descent is $O(\frac{1}{\epsilon})$, which means the number of evaluations (of the gradients) is of the order $O(\frac{1}{\epsilon})$ if we want to obtain an approximate solution with less than $\epsilon$ value error to the optimal solution~\cite{nesterov2003introductory}. This complexity is agnostic to the dimension of parameters to optimize (e.g., $\theta \in \bbR^d$). For a simple zero-order optimization method (such as the simple ES used in this work), the required number of evaluations is $O(\frac{d}{\epsilon})$, which suffers from a factor of $d$. With advanced methods, the overhead factor can be reduced from $d$ to $\sqrt{d}$~\cite{duchi2015optimal, nesterov2017random}. In EPSL, due to the hierarchical structure as shown in Figure~\ref{fig_model_EPSL}, it only needs to estimate the gradient with respect to the decision variable $\vx \in \bbR^n$ rather than the whole model parameters $\vtheta \in \bbR^d$ where $n \ll d$ (e.g., $n = 10$ and $d = 10k$). As reported in Table~\ref{table_runtime}, EPSL typically only requires a comparable computational budget (e.g., number of evaluations and wall-clock runtimes) as in a single run of MOEA/D, while it can learn to approximate the whole Pareto set. 

\end{itemize}

\begin{table}[t]
\caption{Average run time for optimization and sampling on problems with different numbers of objectives.}
\begin{tabular}{cc|c|ccc}
\toprule
     & \# Eval. & MOEA/D & EPSL  & 100 Samples & 1,000 Samples \\ \midrule
RE21 & 25,000   & 16.4s  & 14.3s & +0.00s      & +0.01s        \\
RE31 & 40,000   & 26.2s  & 19.7s & +0.00s      & +0.01s        \\
RE41 & 40,000   & 30.1s  & 24.2s & +0.00s      & +0.01s        \\
RE61 & 40,000   & 28.7s  & 24.7s & +0.01s      & +0.03s        \\
RE91 & 40,000   & 33.8s  & 30.8s & +0.01s      & +0.05s        \\ \bottomrule
\end{tabular}
\label{table_runtime}
\end{table}

\section{Experimental Studies}
\label{sec_experiment}

\begin{figure*}[t]
\centering
\subfloat[Finite Solutions by MOEA/D\label{fig_moea_epsl_a}]{\includegraphics[width = 0.25\linewidth]{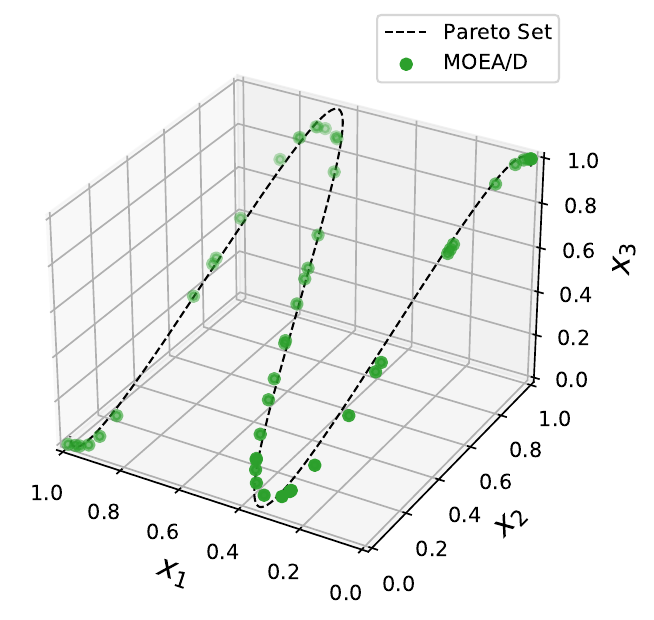}}
\hfill
\subfloat[Learned PS by EPSL\label{fig_moea_epsl_b}]{\includegraphics[width = 0.25\linewidth]{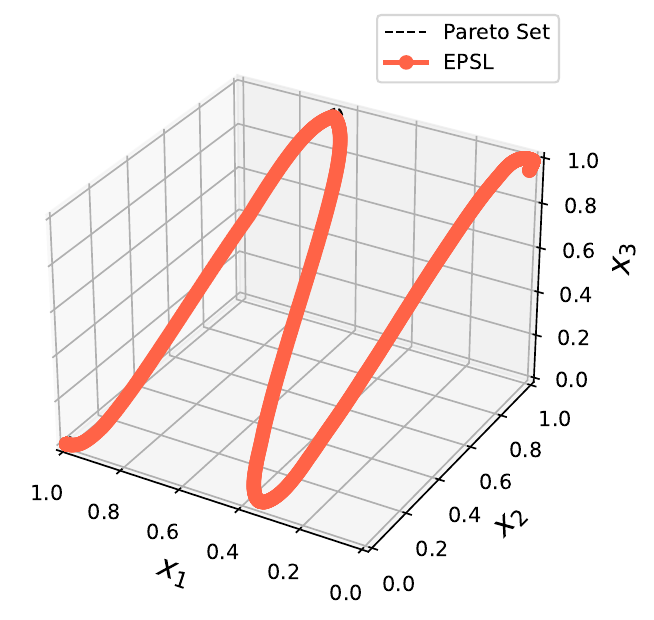}}
\hfill
\subfloat[PS with Structure Constraints\label{fig_moea_epsl_c}]{\includegraphics[width = 0.25\linewidth]{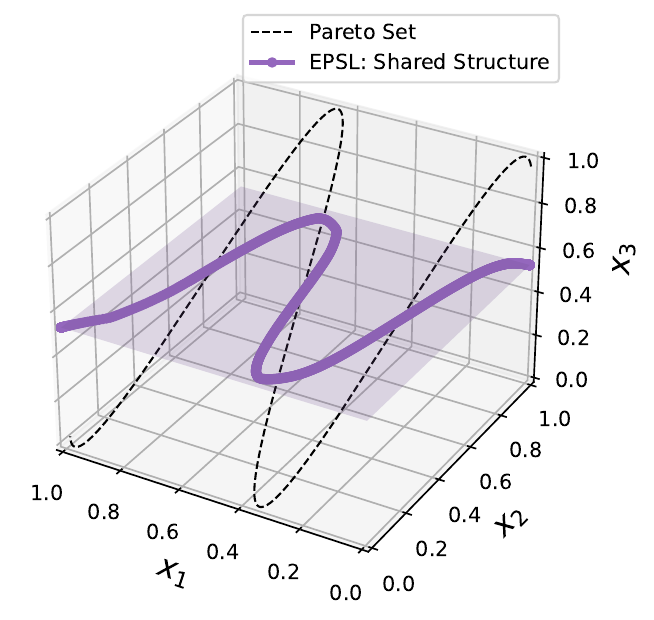}}
\hfill
\subfloat[Learned PF by EPSL\label{fig_moea_epsl_d}]{\includegraphics[width = 0.25\linewidth]{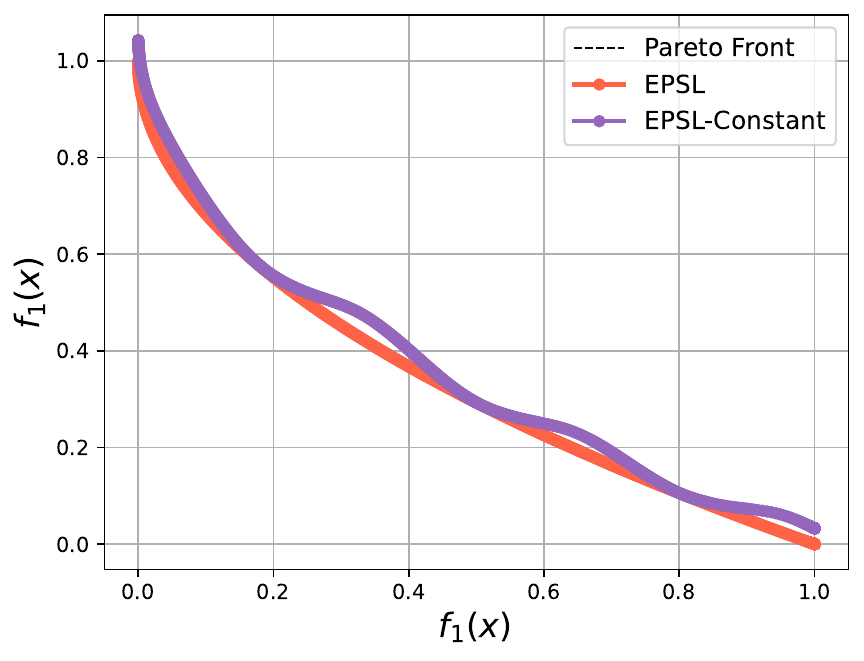}}
\caption{\textbf{Finite Set of Solutions and the Learned Pareto Set:} \textbf{(a)} MOEA/D can only find a finite set of solutions to approximate the Pareto set. \textbf{(b)} The proposed EPSL can successfully approximate the whole Pareto set that contains infinite number of solutions. \textbf{(c)} EPSL can easily incorporate structure constraints on the learned solution set. Here all solutions on the constrained solution set share the same $x_3$. \textbf{(d)} The corresponding Pareto fronts in the objective space. Decision-makers can directly obtain any preferred solutions in real time by exploring the learned Pareto set without further optimization step. For this problem, the learned Pareto front with a shared solution structure is only slightly outperformed by the unconstrained Pareto front.}
\label{fig_moea_epsl}
\end{figure*}

In this section, we conduct two sets of experiments to:
\begin{itemize}

    \item Show EPSL can achieve promising performance, compared with classical MOEAs, on $16$ real-world multiobjective engineering design problems.
    
    \item Demonstrate EPSL can efficiently deal with different structure constraints on the whole solution set for various problems.
    
\end{itemize}

\subsection{Experimental Settings}

\begin{table*}[t]
\small
\center
\caption{Performance of EPSL with different model structures. We use a two-layer fully-connected neural network with 1024 nodes (*) as our Pareto set model in other experiments.}
\begin{tabular}{l|ccc|ccc|ccc}
\hline
                           & \multicolumn{3}{c|}{Single Hidden Layer} & \multicolumn{3}{c|}{Two Hidden Layers}                                                           & \multicolumn{3}{c}{Three Hidden Layers}                                                             \\
Size                 & 128          & 1024        & 2048        & 128-128  & 1024-1024 (*)                     & 2048-2048                                 & 128-128-128 & 512-512-512                               & 1024-1024-1024                            \\ \hline
RE21 & 1.05e-02     & 1.40e-03    & 8.97e-04    & 2.51e-03 & \cellcolor[HTML]{C0C0C0}\textbf{3.91e-04} & \cellcolor[HTML]{C0C0C0}\textbf{3.84e-04} & 1.29e-03    & \cellcolor[HTML]{C0C0C0}\textbf{3.53e-04} & \cellcolor[HTML]{C0C0C0}\textbf{3.37e-04} \\
RE32    & 1.65e-04     & 4.36e-05    & 1.15e-05    & 4.54e-05 & \cellcolor[HTML]{C0C0C0}\textbf{4.72e-06} & \cellcolor[HTML]{C0C0C0}\textbf{4.75e-06} & 2.09e-05    & \cellcolor[HTML]{C0C0C0}\textbf{4.36e-06} & \cellcolor[HTML]{C0C0C0}\textbf{4.65e-06} \\ \hline
\end{tabular}

\label{table_model_sizes}
\end{table*}

\begin{table*}[t!]
\centering
\caption{The median hypervolume gap ($\Delta$HV$\downarrow$) of $21$ independent runs on $16$ RE problems. The $(+/=/-)$ symbols indicate EPSL is significantly better than/equal to/worse than the corresponding algorithm. We report the performance scores in the parenthesis, and highlight the best results in bold and grey background.}
\label{table_re_hv_gap}
\begin{tabular}{lcccccc}
\toprule
\multicolumn{1}{c}{}                                     & NSGA-II                                         & NSGA-III                                        & MOEA/D-TCH                                      & MOEA/D-PBI     & EPSL                                            & EPSL-Large Set                                 \\ \midrule
RE21-Four Bar Truss Design                               & 7.46e-03 (5) +                                  & 5.02e-03 (2) +                                  & 4.93e-03 (1) +                                  & 7.01e-03 (4) + & 5.52e-03 (3) +                                  & \cellcolor[HTML]{C0C0C0}\textbf{3.91e-04 (0)}  \\
\multicolumn{1}{c}{RE22-Reinforced Concrete Beam Design} & 7.32e-03 (3) +                                  & 6.10e-03 (1) -                                  & \cellcolor[HTML]{C0C0C0}\textbf{5.72e-03 (0) -} & 8.84e-03 (4) + & 9.63e-03 (5) +                                  & 6.47e-03 (2)                                   \\
RE23-Pressure Vessel Design                              & \cellcolor[HTML]{C0C0C0}\textbf{1.32e-03 (0) -} & 2.51e-03 (2) +                                  & 4.09e-03 (4) +                                  & 4.78e-03 (5) + & 3.38e-03 (3) +                                  & 1.42e-03 (1)                                   \\
RE24-Hatch Cover Design                                  & 2.39e-03 (2) +                                  & 3.79e-03 (3) +                                  & 3.77e-03 (3) +                                  & 1.16e-02 (5) + & 1.56e-03 (1) +                                  & \cellcolor[HTML]{C0C0C0}\textbf{7.07e-05 (0)}  \\
RE25-Coil Compression Spring Design                      & 5.59e-04 (2) +                                  & 6.24e-03 (3) +                                  & 6.18e-03 (3) +                                  & 1.68e-02 (5) + & \cellcolor[HTML]{C0C0C0}\textbf{6.98e-08 (0) =} & \cellcolor[HTML]{C0C0C0}\textbf{6.98e-08 (0)}  \\
RE31-Two Bar Truss Design                                & 1.01e-04 (2) -                                  & \cellcolor[HTML]{C0C0C0}\textbf{1.04e-05 (0) -} & 2.35e-05 (1) -                                  & 9.51e-03 (5) + & 4.35e-04 (4) +                                  & 4.03e-04 (3)                                   \\
RE32-Welded Beam Design                                  & 2.83e-03 (2) +                                  & 1.19e-02 (3) +                                  & 1.18e-02 (3) +                                  & 3.48e-02 (5) + & 8.30e-04 (1) +                                  & \cellcolor[HTML]{C0C0C0}\textbf{4.72e-06 (0)}  \\
RE33-Disc Brake Design                                   & 1.39e-01 (5) +                                  & 2.96e-02 (2) +                                  & 5.52e-02 (3) +                                  & 7.02e-02 (4) + & 1.89e-02 (1) +                                  & \cellcolor[HTML]{C0C0C0}\textbf{2.65e-04 (0)}  \\
RE34-Vehicle Crashworthiness Design                      & \cellcolor[HTML]{C0C0C0}\textbf{2.42e-02 (0) -} & 3.83e-02 (1) -                                  & 4.28e-02 (2) -                                  & 6.36e-02 (3) - & 7.55e-02 (5) +                                  & 6.56e-02 (4)                                   \\
RE35-Speed Reducer Design                                & 9.89e-03 (3) +                                  & 1.10e-02 (4) +                                  & 4.96e-03 (1) +                                  & 1.02e-02 (4) + & 4.89e-03 (1) +                                  & \cellcolor[HTML]{C0C0C0}\textbf{2.88e-03 (0)}  \\
RE36-Gear Train Design                                   & 3.25e-04 (2) +                                  & 1.74e-02 (5) +                                  & 7.86e-04 (3) +                                  & 8.95e-03 (4) + & -5.48e-03 (1) +                                 & \cellcolor[HTML]{C0C0C0}\textbf{-1.32e-02 (0)} \\
RE37-Rocket Injector Design                              & 6.00e-02 (4) +                                  & 5.68e-02 (4) +                                  & 3.91e-02 (1) +                                  & 4.45e-02 (3) + & 4.04e-02 (2) +                                  & \cellcolor[HTML]{C0C0C0}\textbf{9.27e-03 (0)}  \\
RE41-Car Side Impact Design                              & 1.42e-01 (4) +                                  & 1.38e-01 (4) +                                  & 9.71e-02 (2) +                                  & 8.89e-02 (1) + & 1.06e-01 (3) +                                  & \cellcolor[HTML]{C0C0C0}\textbf{4.65e-02 (0)}  \\
RE42-Conceptual Marine Design                            & 1.66e-02 (3) +                                  & 6.67e-02 (5) +                                  & 4.38e-03 (2) +                                  & 3.23e-02 (4) + & 2.07e-03 (1) +                                  & \cellcolor[HTML]{C0C0C0}\textbf{-1.25e-02 (0)} \\
RE61-Water Resource Planning                             & 1.16e-01 (3) +                                  & 1.42e-01 (5) +                                  & 9.44e-02 (2) +                                  & 1.31e-01 (4) + & 5.75e-03 (1) +                                  & \cellcolor[HTML]{C0C0C0}\textbf{-2.96e-02 (0)} \\
RE91-Car Cab Design                                      & 3.20e-02 (0) -                                  & 3.62e-02 (1) -                                  & 4.37e-02 (2) -                                  & 6.61e-02 (5) + & 4.75e-02 (3) =                                  & 4.75e-02 (3)                                   \\ \midrule
\multicolumn{1}{c}{+/=/-}                                & 12/0/4                                          & 12/0/4                                          & 12/0/4                                          & 15/0/1         & 14/2/0                                          & -                                              \\ \midrule
Average Score                                            & 2.500                                           & 2.8125                                          & 2.0625                                          & 4.0625         & 2.1875                                          & 0.8125                                         \\ \bottomrule
\end{tabular}

\centering
\subfloat[RE21-Four Bar Truss]{\includegraphics[width = 0.235\linewidth]{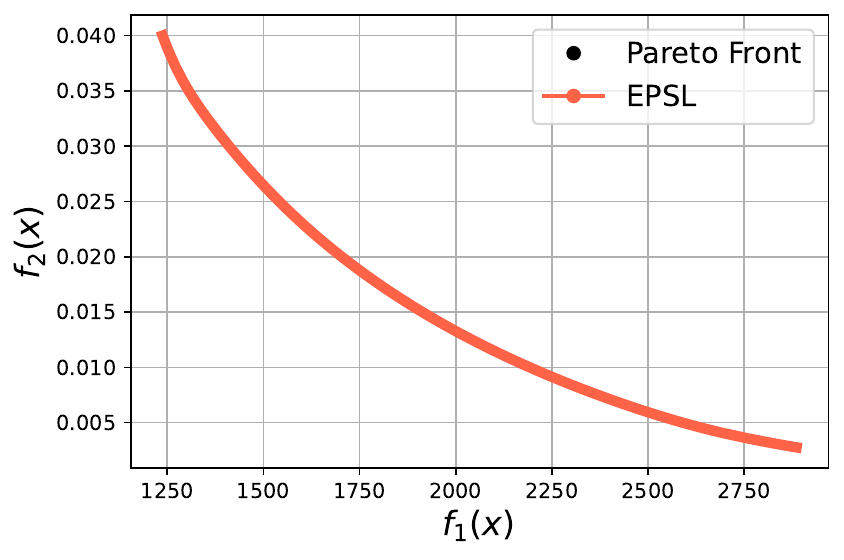}}
\subfloat[RE22-Reinforced Concrete Beam]{\includegraphics[width = 0.235\linewidth]{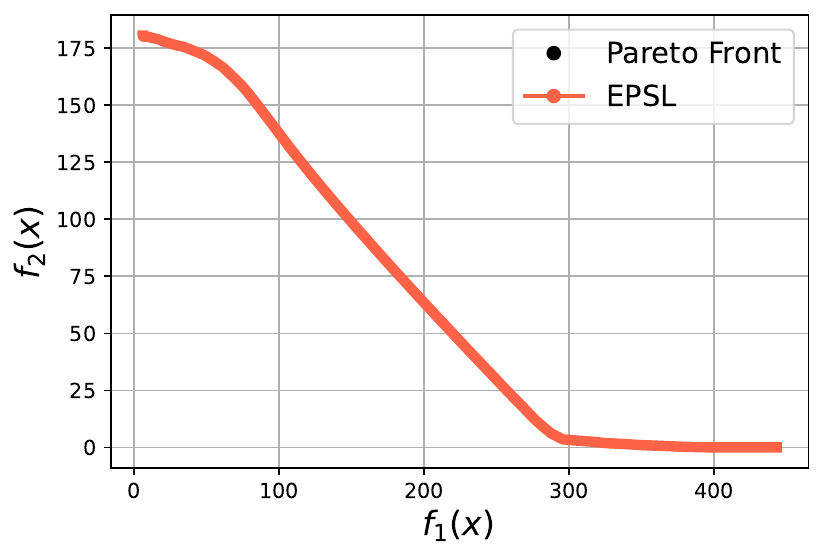}}
\subfloat[RE23-Pressure Vessel]{\includegraphics[width = 0.235\linewidth]{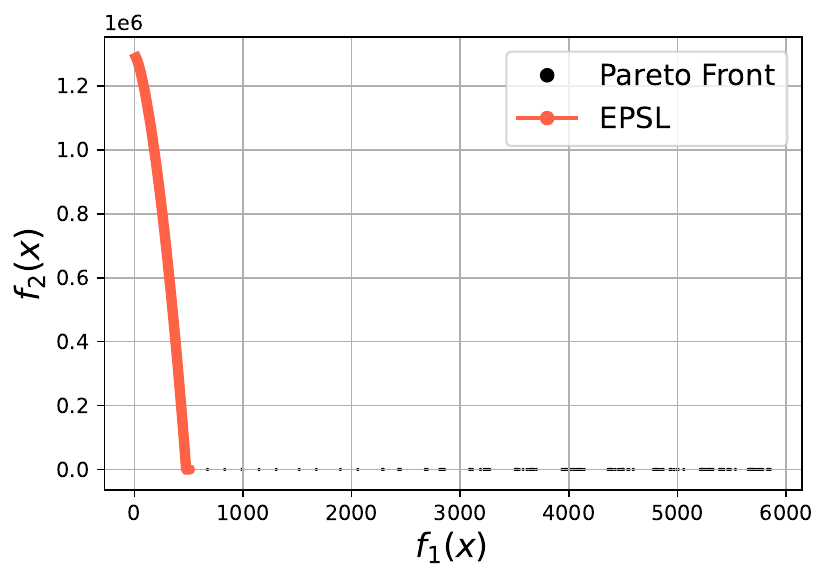}}
\subfloat[RE24-Hatch Cover]{\includegraphics[width = 0.235\linewidth]{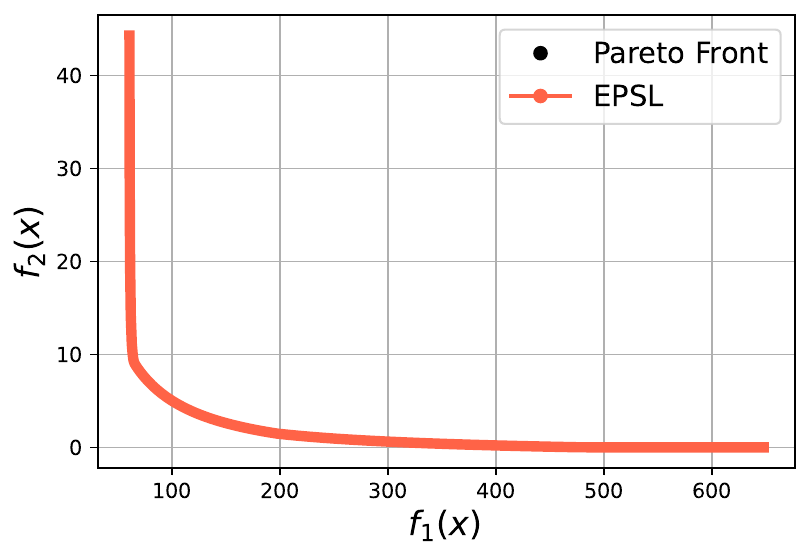}} \\
\subfloat[RE32-Welded Beam Design]{\includegraphics[width = 0.235\linewidth]{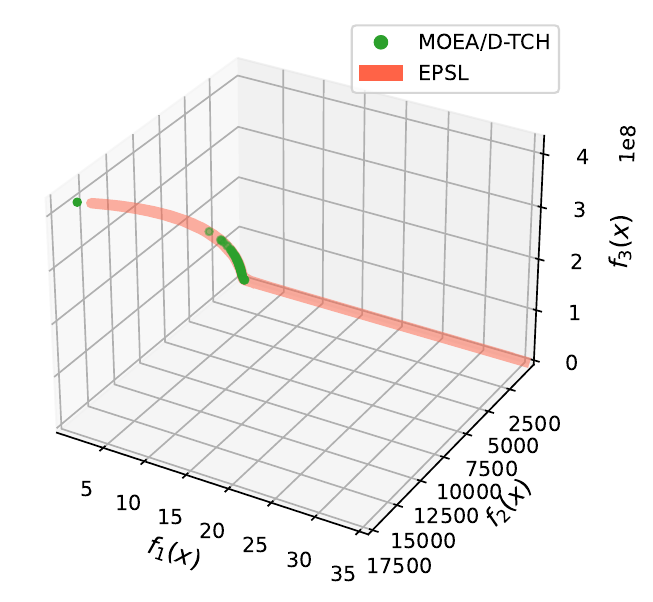}}
\subfloat[RE33-Disc Brake Design]{\includegraphics[width = 0.235\linewidth]{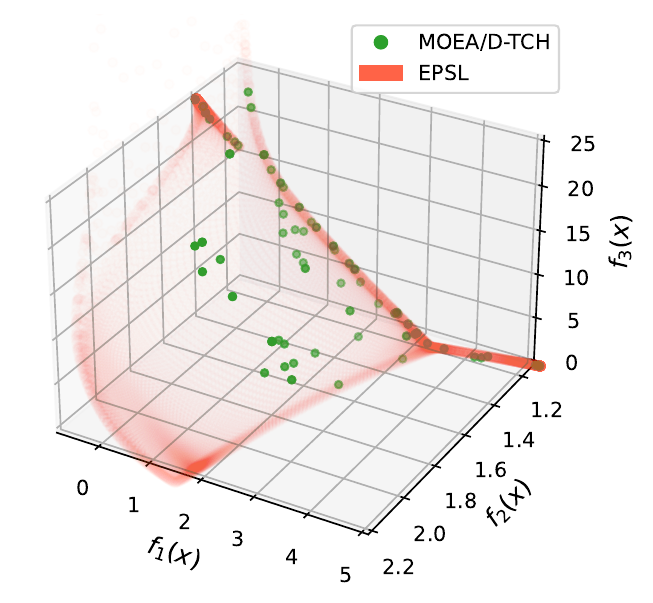}}
\subfloat[RE36-Gear Train Design]{\includegraphics[width = 0.235\linewidth]{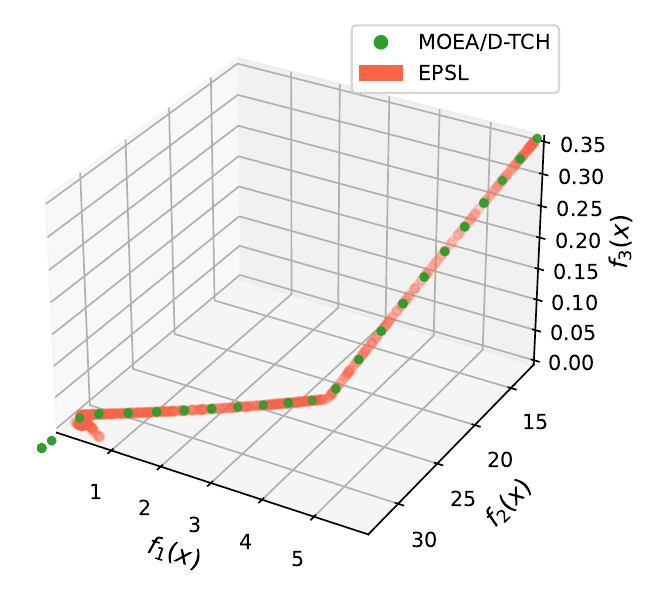}}
\subfloat[RE37-Rocket Injector Design]{\includegraphics[width = 0.235\linewidth]{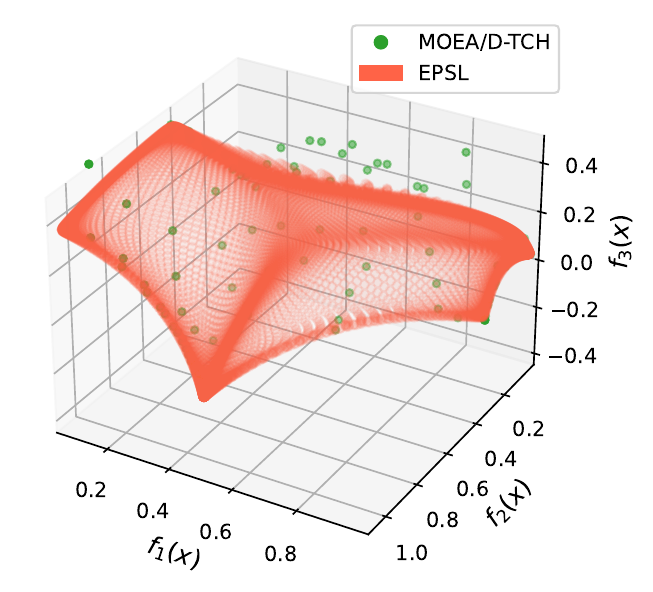}} \\
\subfloat[RE32-Welded Beam Design]{\includegraphics[width = 0.235\linewidth]{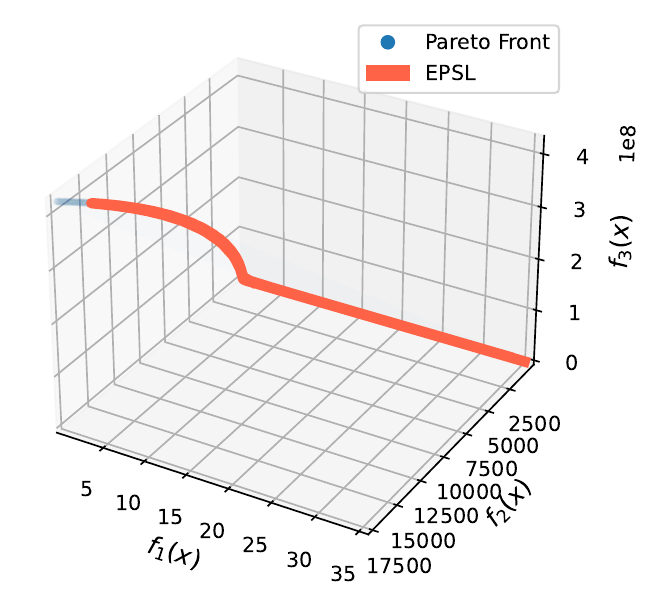}}
\subfloat[RE33-Disc Brake Design]{\includegraphics[width = 0.235\linewidth]{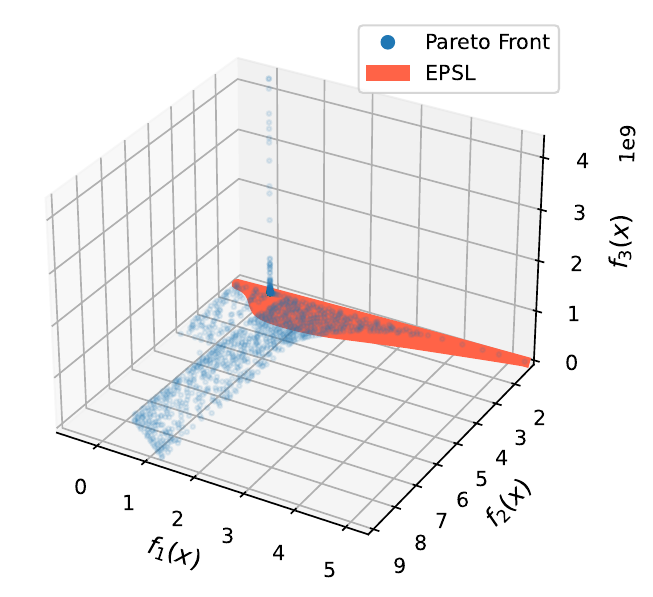}}
\subfloat[RE36-Gear Train Design]{\includegraphics[width = 0.235\linewidth]{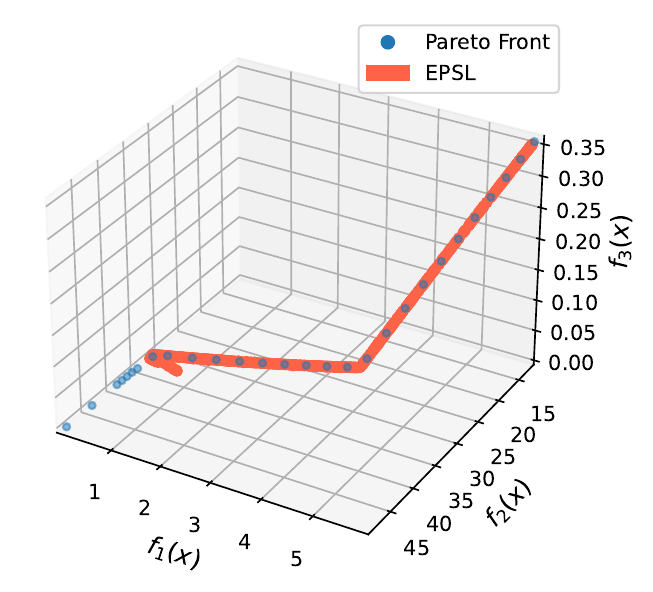}}
\subfloat[RE37-Rocket Injector Design]{\includegraphics[width = 0.235\linewidth]{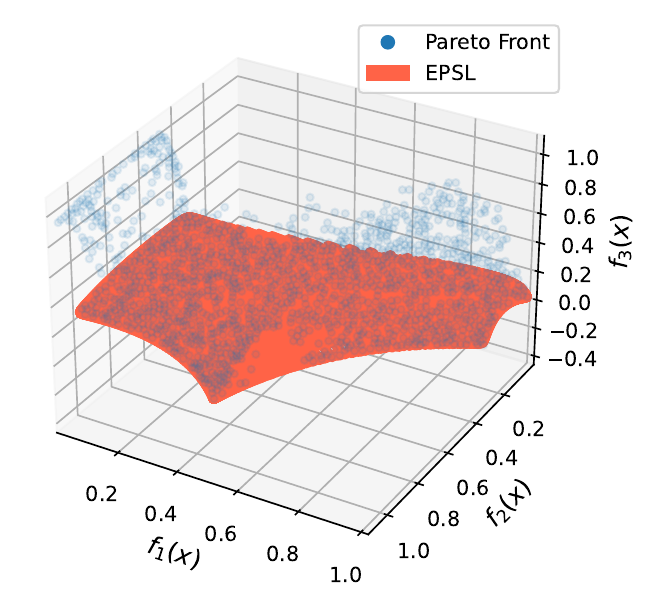}}
\captionof{figure}{\textbf{The Learned Pareto Front for RE Problems:} \textbf{(a)-(d)} EPSL is able to learn the Pareto front for bi-objective problems. \textbf{(e)-(h)} For the three objective optimization problems, EPSL allows the decision-makers to easily explore different trade-off solutions on the learned Pareto front, while MOEA/D can only provide a finite set of solutions. \textbf{(i)-(l)} Compared with the approximated Pareto front with exhausted search by different methods, EPSL can learn the part of Pareto front with critical trade-offs, but it might fail to find the regions which mainly contain weakly Pareto efficient solutions.}
\label{fig_results_re2}
\end{table*}

\subsubsection{Optimization Problems}

The multiobjective engineering design optimization test instance suite (RE problems) proposed in~\cite{tanabe2020easy}\footnote{\href{https://github.com/ryojitanabe/reproblems}{https://github.com/ryojitanabe/reproblems}} are used in our experimental studies.
This instance suite contains $16$ different real-world engineering design problems with $2,3,4,6$ and $9$ optimization objectives from various domains such as bar truss design~\cite{coello2005multiobjective} and car cab design~\cite{deb2013evolutionary}. These problems have different properties and shapes of the Pareto front. A detailed summary of these problems can be found in \cite{tanabe2020easy}.

\subsubsection{Algorithm Setting} Our proposed EPSL method has a strong connection to the decomposition-based algorithms. In this section, we compare it with three representative decomposition-based and preference-based MOEA, namely MOEA/D~\cite{zhang2007moea} with Tchebycheff (MOEA/D-TCH) and PBI decomposition (MOEA/D-PBI), NSGA-III~\cite{deb2013evolutionary}, as well as the seminal dominance-based NSGA-II~\cite{deb2002fast}. We implement all MOEAs with the open-sourced pymoo~\cite{pymoo}\footnote{\href{https://pymoo.org/index.html}{https://pymoo.org/index.html}} and our proposed EPSL\footnote{Will be open-sourced upon publication.} with PyTorch~\cite{paszke2019pytorch}.

All methods are given the same number of function evaluations for each problem, which is $26,000$ for two objective problems, and $41,000$ for problems with three or more objectives. Similar to \cite{tanabe2020easy}, all MOEAs have the population size $100, 105, 120, 182, 90$ for problems with $2, 3, 4, 6, 9$ objectives, respectively. The evolutionary operators are SBX crossover and polynomial mutation for all MOEAs, and we use the default settings for other parameters as in their original papers.

Our proposed EPSL is trained by the lightweight evolutionary stochastic gradient optimization method introduced in Section~\ref{subsec_esgd}. It randomly samples $5$ (for two objective problems) or $8$ (for other problems) valid preferences at each iteration, and samples $5$ solutions to estimate the gradient for each preference. The total number of iterations is $1,000$ for all problems. At the end of optimization, we randomly sample $[100, 105, 120, 182, 90]$ (match those for MOEAs) or $[1000, 990, 969, 923, 90]$ (a larger set) preferences for the problems with $2, 3, 4, 6, 9$ objectives, and obtain their corresponding Pareto solutions for comparison with other MOEAs. The number of preferences for the 9-objective problem is still $90$ in the larger set to avoid the extremely long hypervolume calculation. Therefore, the total number of evaluations is at most $26,000$ or $41,000$ for problems with 2 or more objectives, respectively. The average running times for problems with different numbers of objectives are reported in Table~\ref{table_runtime}. For a fair comparison, all algorithms are run on the same CPU. With the proposed efficient stochastic evolutionary gradient descent algorithm, EPSL is comparable with MOEA/D in terms of running time on all test instances. In addition, sampling solutions from the learned Pareto set is trivial and their computational cost can be neglected.

\begin{table*}[t]

\center
\caption{Performance of EPSL with weighted-sum (WS), penalty boundary intersection (PBI), and Tchebycheff (TCH) aggregation methods. We report both the results with $100$ and $1,000$ samples.}

\begin{tabular}{lcccc}
\toprule
                                                         & \multicolumn{4}{c}{EPSL-Large Set}                                                                                                                                    \\
\multicolumn{1}{c}{}                                     & WS              & PBI                                              & TCH                                             & STCH                                           \\ \midrule
RE21-Four Bar Truss Design                               & 3.02e-04 (1) -  & \cellcolor[HTML]{C0C0C0}\textbf{1.52e-04 (0) -}  & 1.88e-03 (3) +                                  & 3.91e-04 (0)                                   \\
\multicolumn{1}{c}{RE22-Reinforced Concrete Beam Design} & 1.94e-02 (1) +  & 2.07e-02 (2) +                                   & 2.32e-02 (0) +                                  & \cellcolor[HTML]{C0C0C0}\textbf{6.47e-03 (2)}  \\
RE23-Pressure Vessel Design                              & 3.86e-03 (3) +  & 3.65e-03 (2) +                                   & 3.26e-03 (1) +                                  & \cellcolor[HTML]{C0C0C0}\textbf{1.42e-03 (1)}  \\
RE24-Hatch Cover Design                                  & 6.52e-04 (2) +  & 1.43e-04 (1) +                                   & 9.09e-04 (3) +                                  & \cellcolor[HTML]{C0C0C0}\textbf{7.07e-05 (0)}  \\
RE25-Coil Compression Spring Design                      & 7.44e-08 (3) +  & \cellcolor[HTML]{C0C0C0}\textbf{6.98e-08 (0) =}  & \cellcolor[HTML]{C0C0C0}\textbf{6.98e-08 (0) =} & \cellcolor[HTML]{C0C0C0}\textbf{6.98e-08 (0)}  \\
RE31-Two Bar Truss Design                                & 1.08e-02 (3) +  & 5.13e-03 (2) +                                   & 3.03e-04 (1) +                                  & \cellcolor[HTML]{C0C0C0}\textbf{4.03e-04 (3)}  \\
RE32-Welded Beam Design                                  & 1.14e-04 (1) +  & 3.06e-04 (3) +                                   & 1.84e-04 (2) +                                  & \cellcolor[HTML]{C0C0C0}\textbf{4.72e-06 (0)}  \\
RE33-Disc Brake Design                                   & 2.30e-02 (3) +  & 1.75e-03 (2) +                                   & 1.60e-03 (1) +                                  & \cellcolor[HTML]{C0C0C0}\textbf{2.65e-04 (0)}  \\
RE34-Vehicle Crashworthiness Design                      & 2.82e-01 (3) +  & 7.93e-02 (1) +                                   & 1.42e-01 (2) +                                  & \cellcolor[HTML]{C0C0C0}\textbf{6.56e-02 (4)}  \\
RE35-Speed Reducer Design                                & 5.70e-02 (3) +  & 5.32e-02 (2) +                                   & 7.97e-03 (1) +                                  & \cellcolor[HTML]{C0C0C0}\textbf{2.88e-03 (0)}  \\
RE36-Gear Train Design                                   & -1.20e-02 (3) + & -1.25e-02 (2) +                                  & -1.28e-02 (1) +                                 & \cellcolor[HTML]{C0C0C0}\textbf{-1.32e-02 (0)} \\
RE37-Rocket Injector Design                              & 1.33e-01 (3) +  & 1.76e-02 (2) +                                   & 1.29e-02 (1) +                                  & \cellcolor[HTML]{C0C0C0}\textbf{9.27e-03 (0)}  \\
RE41-Car Side Impact Design                              & 9.66e-02 (3) +  & 6.38e-02 (1) +                                   & 7.59e-02 (2) +                                  & \cellcolor[HTML]{C0C0C0}\textbf{4.65e-02 (0)}  \\
RE42-Conceptual Marine Design                            & 4.24e-02 (3) +  & \cellcolor[HTML]{C0C0C0}\textbf{-1.34e-02 (0) -} & -7.45e-03 (2) +                                 & -1.25e-02 (0)                                  \\
RE61-Water Resource Planning                             & -2.35e-02 (3) + & -2.75e-02 (1) +                                  & -2.47e-02 (2) +                                 & \cellcolor[HTML]{C0C0C0}\textbf{-2.96e-02 (0)} \\
RE91-Car Cab Design                                      & 5.33e-02 (3) +  & \cellcolor[HTML]{C0C0C0}\textbf{4.75e-02 (0) =}  & \cellcolor[HTML]{C0C0C0}\textbf{4.75e-02 (0) =} & \cellcolor[HTML]{C0C0C0}\textbf{4.75e-02 (3)}  \\ \midrule
\multicolumn{1}{c}{+/=/-}                                & 15/0/1          & 12/2/2                                           & 14/0/2                                          & -                                              \\ \midrule
\multicolumn{1}{c}{Average Score}                        & 2.5625          & 1.3125                                           & 1.3750                                          & 0.1875                                         \\ \bottomrule
\end{tabular}
\label{table_epsl_aggregation}
\end{table*}

\subsubsection{Performance Indicator}
We use hypervolume (HV)~\cite{zitzler1999multiobjective} as the performance indicator, and report the hypervolume difference ($\Delta$HV) with the approximated Pareto front provided by \cite{tanabe2020easy} for each algorithm. To calculate the hypervolume values, we first normalize all obtained solutions into the unit space $[0,1]^m$ with the approximated ideal point $z^{\text{ideal}}$ and nadir point $z^{\text{nadir}}$ provided in \cite{tanabe2020easy}, and then calculate the hypervolume values with reference vector $(1.1,\cdots, 1.1)^{m}$. We run each algorithm $21$ times for each problem and report the median value. For each comparison, we conduct a Wilcoxon rank-sum test at the $5\%$ significance level and calculate the performance score~\cite{bader2011hype} for each algorithm.

\subsubsection{Model Setting for EPSL}

In this work, we use a simple two-layer fully connected neural network as the Pareto set model, and find it works well for all problems without specific tuning. We have conducted an experimental analysis on our proposed EPSL method with different model structures on the RE21 and RE32 problems as shown in Table.~\ref{table_model_sizes}. The best and close to best results are highlighted in \textbf{bold}. According to the results, a very small model (e.g., those with a single hidden layer) does not have enough capacity to learn the whole Pareto set, and hence has a poor performance. On the other hand, once the model has sufficient capacity, further increasing the model size will not lead to significantly better performance.

\subsection{Illustrative Synthetic Problem}

We first consider the following synthetic two-objective optimization problem to illustrate the properties of EPSL:
\begin{align}
    &\min_{\vx_1 \in [0,1], \vx_{[2:n]} \in [-1,1]^{n-1}} (f_1(\vx),f_2(\vx)), \text{where} \nonumber \\
    &f_1(\vx) = \vx_1, \nonumber \\
    &f_2(\vx) = (1+g)(1 - \sqrt{\vx_1/(1+g)}), \nonumber \\
    &g = \frac{1}{n-1} \sum_{i=2}^{n}(\vx_i - \sin(10(\vx_1 - 0.5)))^2,
\label{synthetic_problem}
\end{align}
of which the Pareto optimal set is a sine-like curve. From the results shown in Fig.~\ref{fig_moea_epsl}(a)(b), it is clear that EPSL can approximate the whole Pareto set while classic MOEA can only find a finite set of solutions. With the learned Pareto set model, decision-makers can easily adjust their preferences and explore the corresponding solutions on the approximate Pareto set in real time without extra optimization steps. In Fig.~\ref{fig_moea_epsl}(c)(d), a structure constraint on the whole solution set is further incorporated into EPSL. We require all solutions should share the same component $x_3$ whose optimal value is unknown in advance. It is clear that EPSL can successfully learn the solution set with such structure constraints.

\begin{figure*}[t]
\centering
\subfloat[Without Constraint]{\includegraphics[width = 0.20\linewidth]{Figures/epsl_re21_pf.pdf}}
\subfloat[Shared $\vx_1$]{\includegraphics[width = 0.20\linewidth]{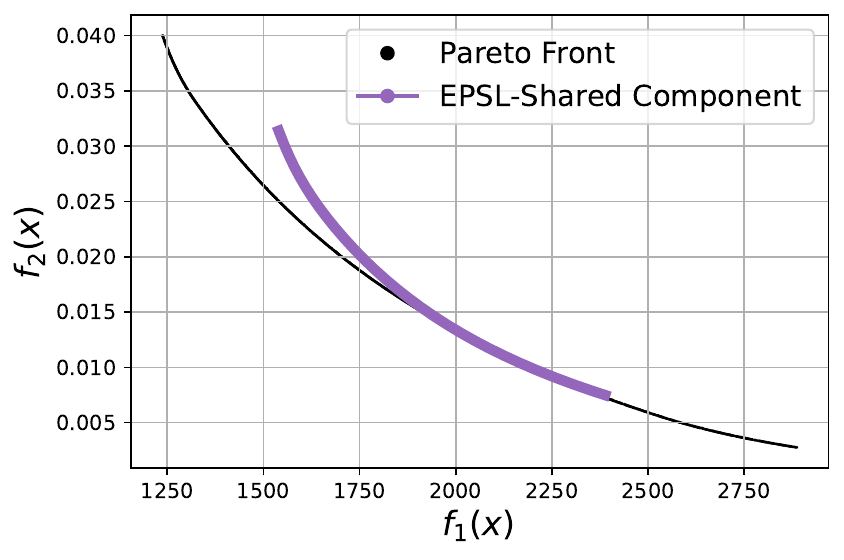}}
\subfloat[Shared $\vx_2$]{\includegraphics[width = 0.20\linewidth]{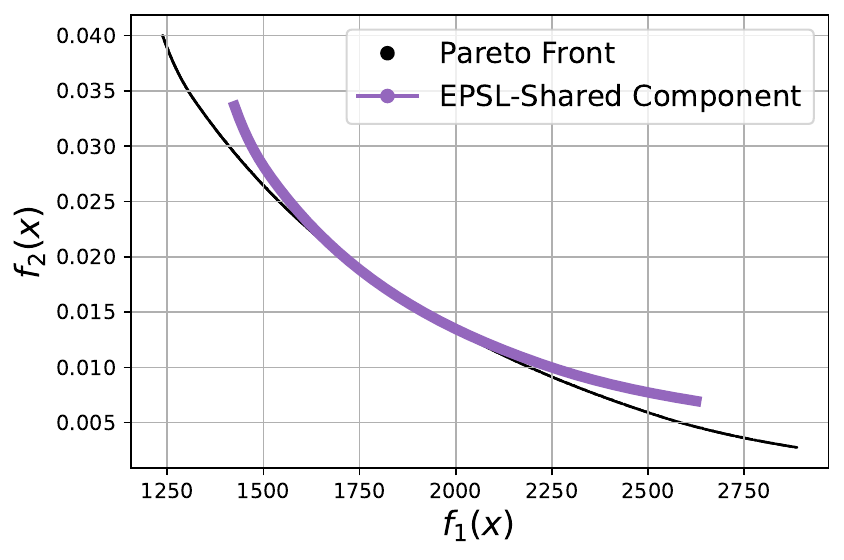}}
\subfloat[Shared $\vx_3$]{\includegraphics[width = 0.20\linewidth]{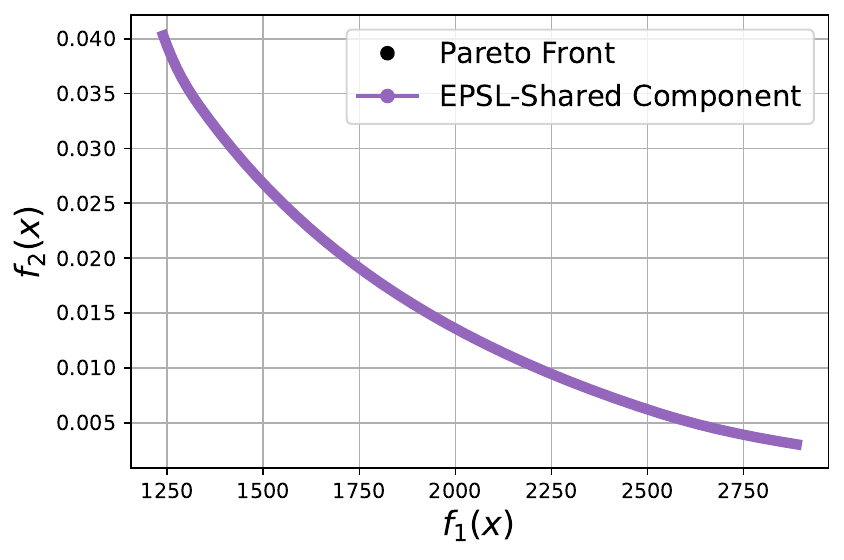}}
\subfloat[Shared $\vx_4$]{\includegraphics[width = 0.20\linewidth]{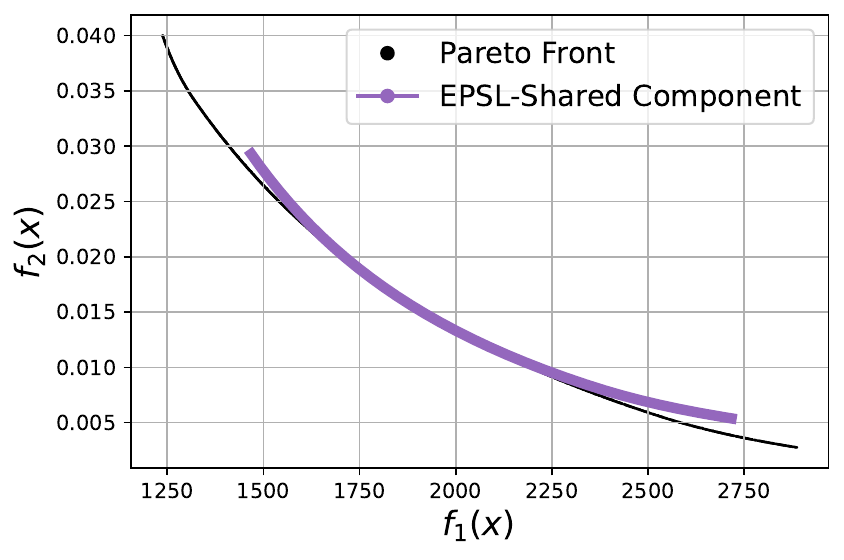}}
\caption{\textbf{The Learned Solution Set for the Four Bar Truss Design Problem (RE21) with Shared Components:} \textbf{(a)} The learned Pareto front without structure constraint. \textbf{(b)-(e)} The image of solution set where the decision variable $\vx_1$, $\vx_2$, $\vx_3$, $\vx_4$ (the lengths of four bars) are shared respectively.}
\label{fig_results_re21_shared_component}
\end{figure*}

\begin{figure*}[t]
\centering
\subfloat[Without Constraint]{\includegraphics[width = 0.20\linewidth]{Figures/epsl_re22_pf.pdf}}
\subfloat[Shared $\vx_1$]{\includegraphics[width = 0.20\linewidth]{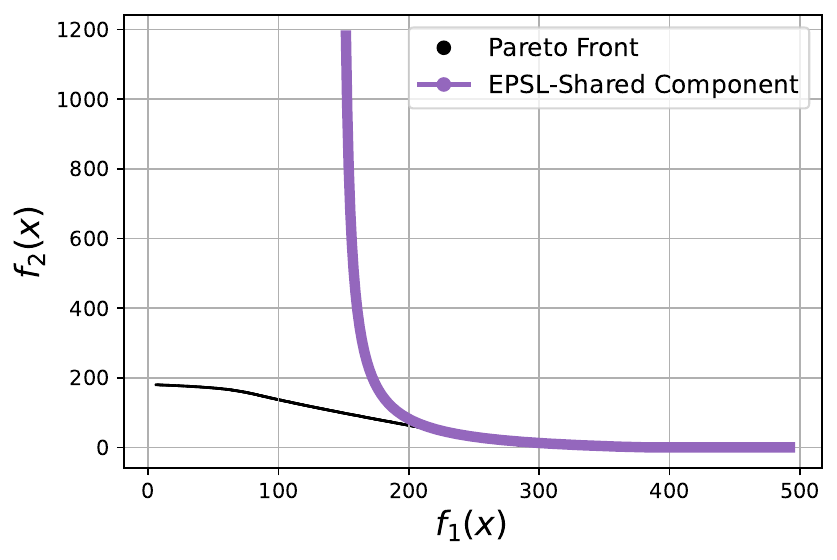}}
\subfloat[Shared $\vx_2$]{\includegraphics[width = 0.20\linewidth]{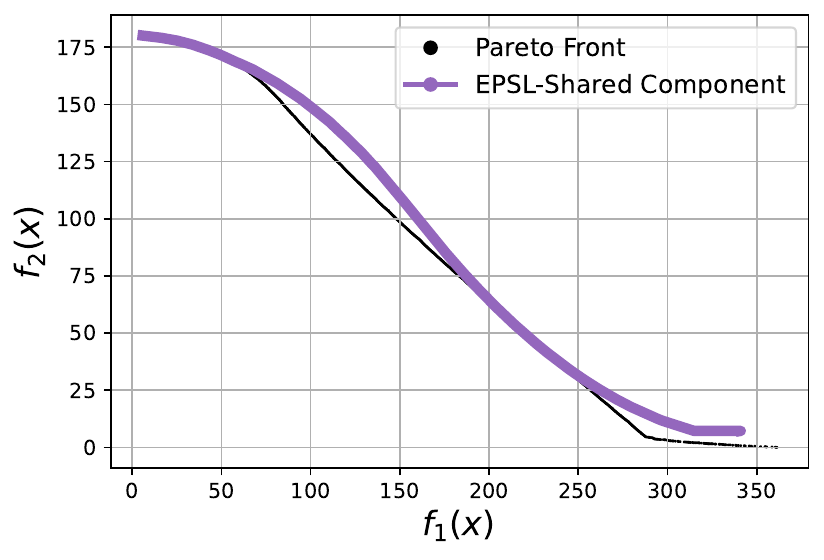}}
\subfloat[Shared $\vx_3$]{\includegraphics[width = 0.20\linewidth]{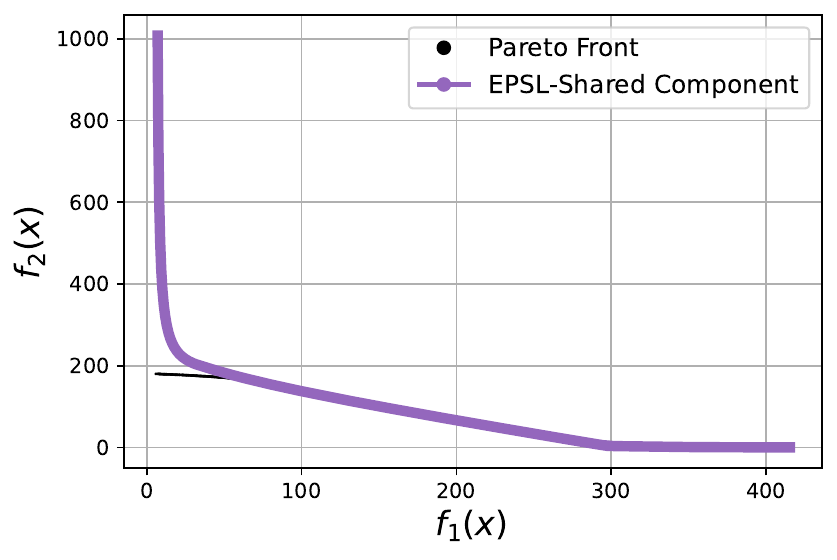}}
\subfloat[Shared $\vx_2$ and $\vx_3$]{\includegraphics[width = 0.20\linewidth]{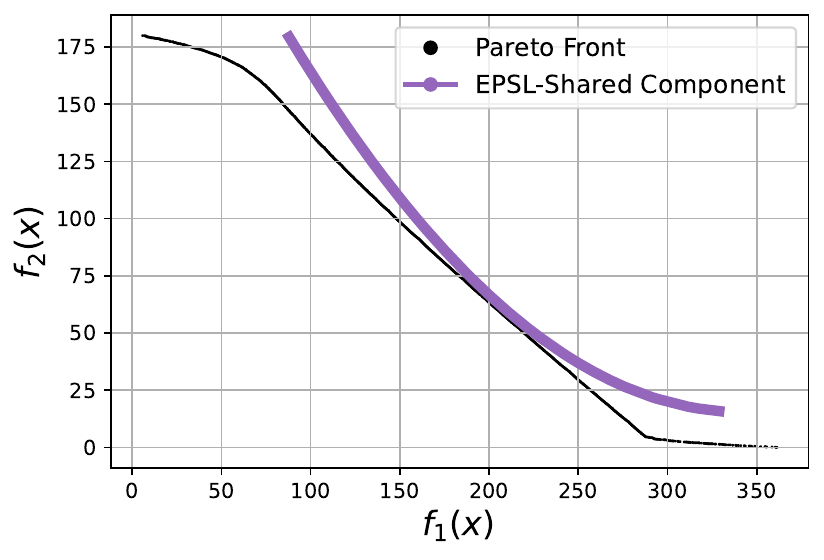}}
\caption{\textbf{The Learned Solution Set for the Reinforced Concrete Beam Design Problem (RE22) with Shared Components:} \textbf{(a)} The learned Pareto front without structure constraint. \textbf{(b)-(e)} The image of learned solution set where the decision variable $\vx_1$ (area of the reinforcement), $\vx_2$ (width of the beam), $\vx_3$ (depth of the beam), or both $\vx_2$ and $\vx_3$ are shared, respectively.}
\label{fig_results_re23_shared_component}
\end{figure*}

\begin{figure*}[t]
\centering
\subfloat[Without Constraint]{\includegraphics[width = 0.20\linewidth]{Figures/re37_epsl_true_pf.pdf}}
\subfloat[Shared $\vx_1$]{\includegraphics[width = 0.20\linewidth]{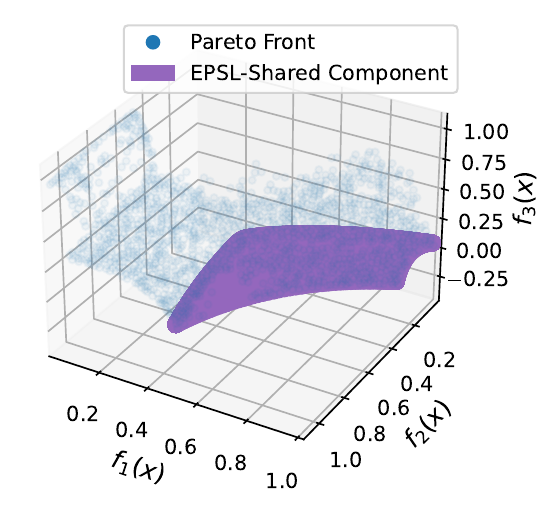}}
\subfloat[Shared $\vx_2$]{\includegraphics[width = 0.20\linewidth]{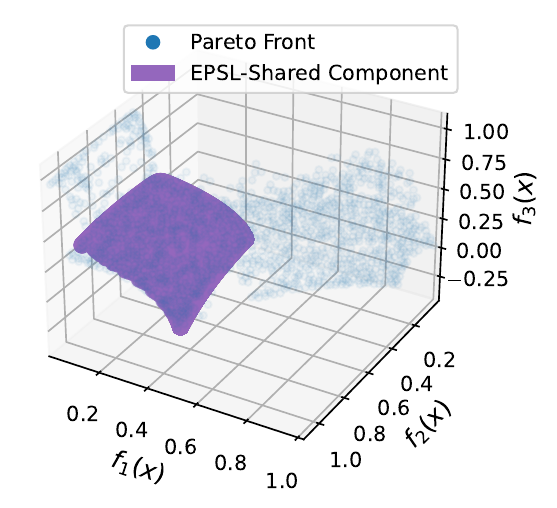}}
\subfloat[Shared $\vx_3$]{\includegraphics[width = 0.20\linewidth]{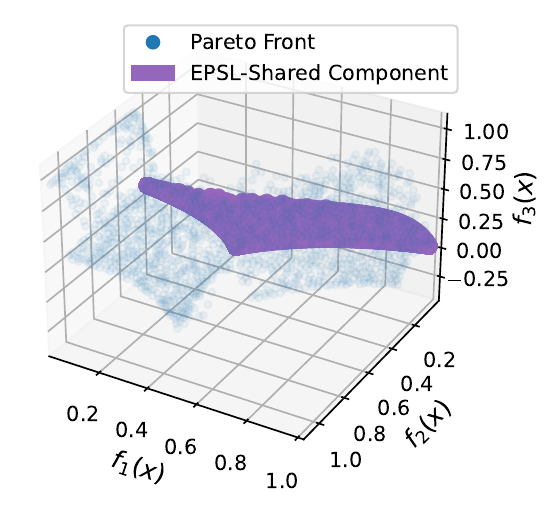}}
\subfloat[Shared $\vx_4$]{\includegraphics[width = 0.20\linewidth]{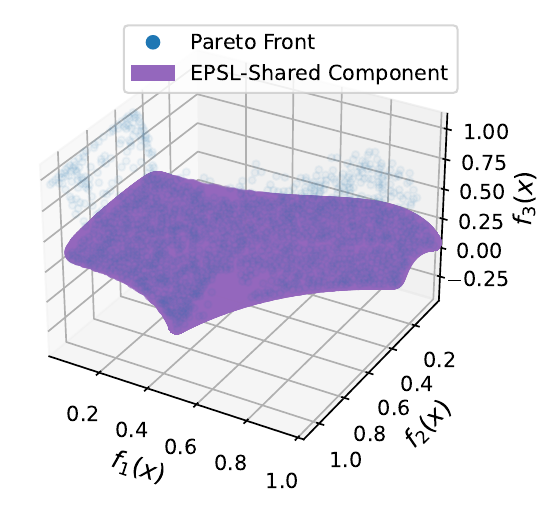}}
\caption{\textbf{The Learned Solution Set for the Rocket Injector Design Problem (RE37) with Shared Components:} \textbf{(a)} The learned Pareto front without structure constraint. \textbf{(b)-(e)} The image of learned solution set where the decision variable $\vx_1$ (hydrogen flow angle), $\vx_2$ (hydrogen area), $\vx_3$ (oxygen area), $\vx_4$ (oxidizer post tip thickness) are shared respectively.}
\label{fig_results_re37_shared_component}
\end{figure*}

\subsection{Unconstrained Pareto Set Learning}

The experimental results on hypervolume difference ($\Delta$HV) for $16$ different real-world multiobjective engineering design problems (RE problems)~\cite{tanabe2020easy} are shown in Table.~\ref{table_re_hv_gap}. Our proposed EPSL method can approximate the whole Pareto set for a given problem via a single model, which is significantly different from a fixed finite set of solutions obtained by other MOEAs. With the learned model, we can randomly sample an arbitrary number of solutions from the approximate Pareto set and calculate its hypervolume difference. In Table.~\ref{table_re_hv_gap}, we report the results of two sampled solution sets by EPSL. The first one (named EPSL) has the same number of solutions with other MOEAs, which is used for a fair comparison. We also report the results for a dense solution set with a larger number of solutions (named EPSL-Large Set) to show the extra advantage of our approach. It should be noticed that the total number of evaluations for EPSL is either smaller than or equal to other MOEAs in comparison. The computational overhead is trivial for the solution sampling as reported in Table.~\ref{table_runtime}.

According to the results in Table.~\ref{table_re_hv_gap}, EPSL performs comparably with other MOEAs with the same number of solutions. By further sampling more solutions on the learned Pareto set, EPSL can obtain the best overall performance. The promising ability to approximate the whole Pareto set makes EPSL a flexible alternative to traditional MOEAs. Some Pareto fronts learned by EPSL are shown in Fig.~\ref{fig_results_re2}. For the two objective problems, EPSL can well approximate the major part of the Pareto front for all the test instances, except the part of weakly Pareto optimal solutions in RE23. For the three objective problems, we compare EPSL with MOEA/D-TCH and the Pareto set provided in \cite{tanabe2020easy}, which is obtained from multiple exhaustive searches of different algorithms with a significantly larger number of evaluations. Compared to MOEA/D-TCH, EPSL can learn a significantly larger part of the Pareto front than a finite set of solutions. On the dense Pareto front, although EPSL will sometimes miss the part of weakly Pareto optimal solutions, it can fully explore the major part of the Pareto front with significant and meaningful trade-offs (with small or even negative $\Delta HV$ in Table.~\ref{table_re_hv_gap}). In this work, we use a simple model and optimization algorithm for EPSL. It is worth investigating more powerful models, sampling methods, and optimization algorithms to further improve the performance of EPSL in future work.

In this work, we mainly use the smooth Tchebycheff (STCH) scalarization for Pareto set learning. We also conduct experiments on EPSL with the weighted-sum (WS), penalty boundary intersection (PBI) and Tchebycheff (TCH) aggregation methods as shown in Table.~\ref{table_epsl_aggregation}. According to the results, EPSL-WS performs poorly on most problems, while EPSL-PBI has a comparable overall performance with EPSL-TCH. EPSL-STCH significantly outperforms other methods on most instances. These results are reasonable since EPSL is a model-based extension and shares many properties similar to MOEA/D. For example, the simple weighted-sum (WS) aggregation cannot find any non-convex parts of the Pareto front and hence will lead to poor overall performance. The nonsmooth nature of TCH and PBI (e.g., \texttt{max} and \texttt{abs} operators) make them suffer from slow convergence with evolutionary stochastic gradient based optimization.

\begin{figure*}[t]
\centering
\subfloat[Learned PS by EPSL]{\includegraphics[width = 0.25\linewidth]{Figures/syn_sin_ps_epsl_new.pdf}}
\hfill
\subfloat[Learned PS with Sine Relation]{\includegraphics[width = 0.25\linewidth]{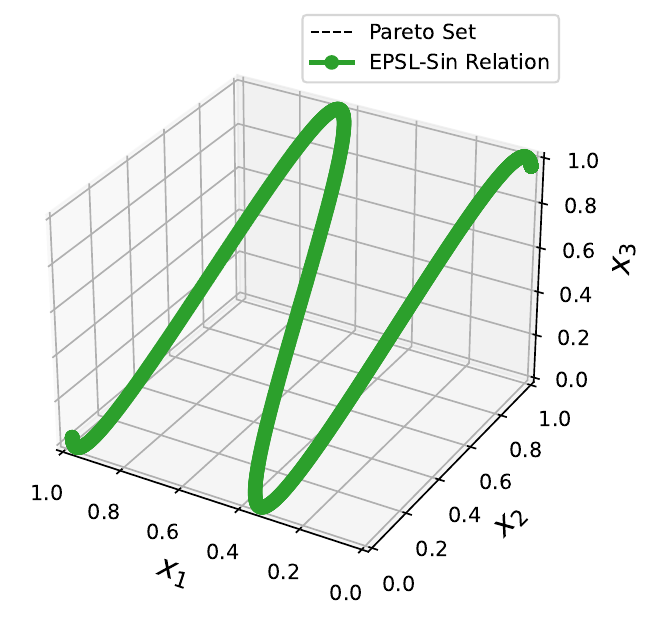}}
\hfill
\subfloat[PF by EPSL]{\includegraphics[width = 0.25\linewidth]{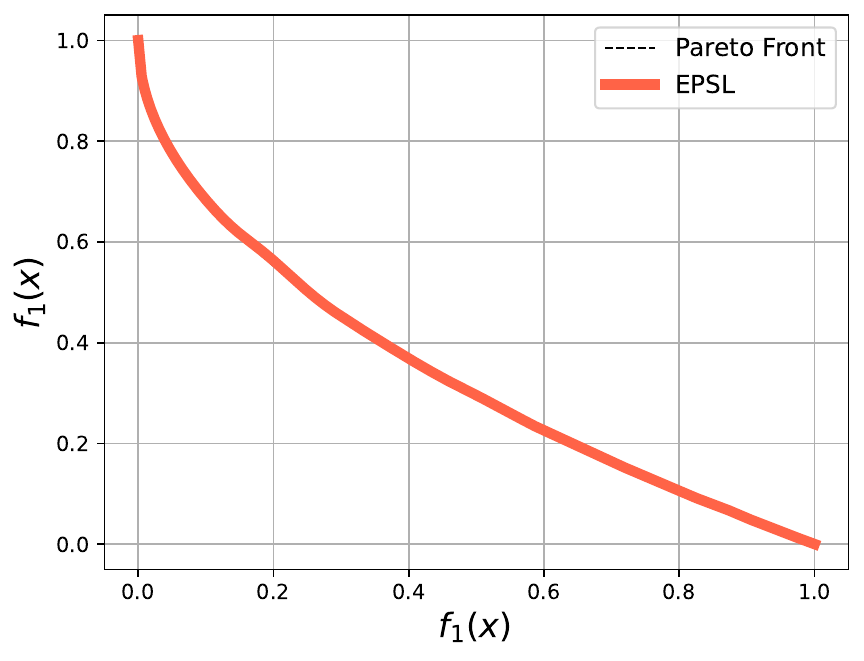}}
\hfill
\subfloat[PF by EPSL with Sine Relation]{\includegraphics[width = 0.25\linewidth]{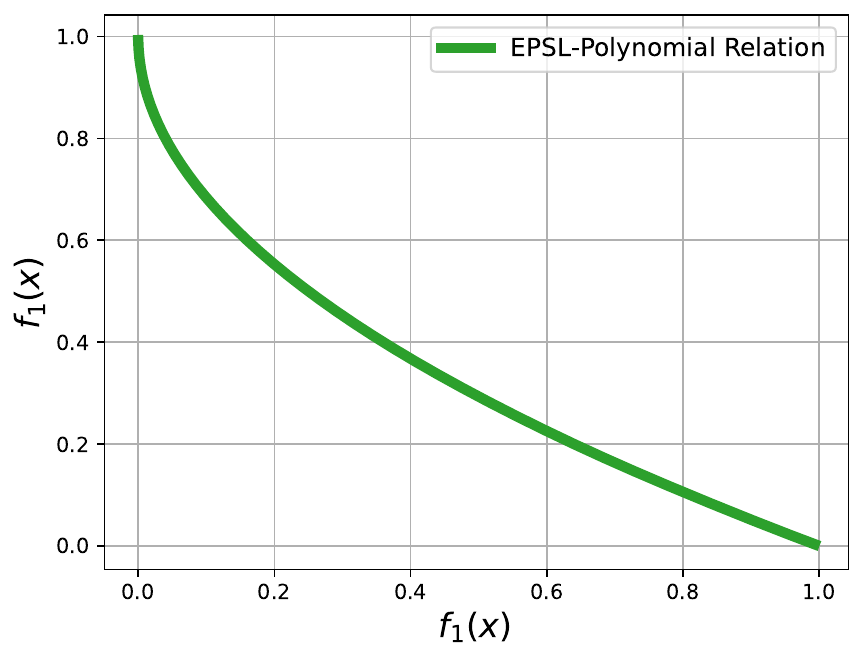}}
\caption{\textbf{EPSL with Learnable Sine-Curve Relationship for the Synthetic Optimization Problem:} \textbf{(a)} The approximate Pareto set learned by the original EPSL. \textbf{(b)} The approximate Pareto set with learnable sine-curve relation. \textbf{(c)} The approximate Pareto front obtained by the original EPSL. \textbf{(d)} The approximate Pareto front corresponds to the learned Pareto set with the sine-curve structure.}
\label{fig_epsl_variable_relation_syn}
\end{figure*}

\begin{figure*}[t]
\centering
\subfloat[Learned PS by EPSL]{\includegraphics[width = 0.25\linewidth]{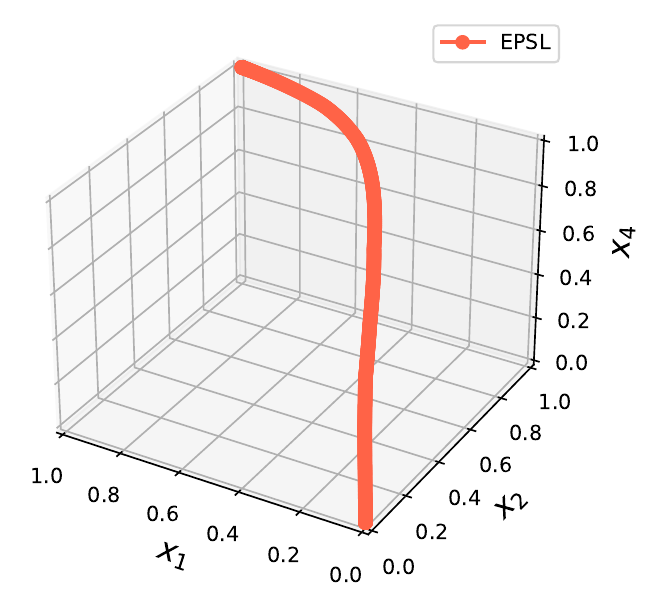}}
\hfill
\subfloat[Learned PS with Polynomial Shape]{\includegraphics[width = 0.25\linewidth]{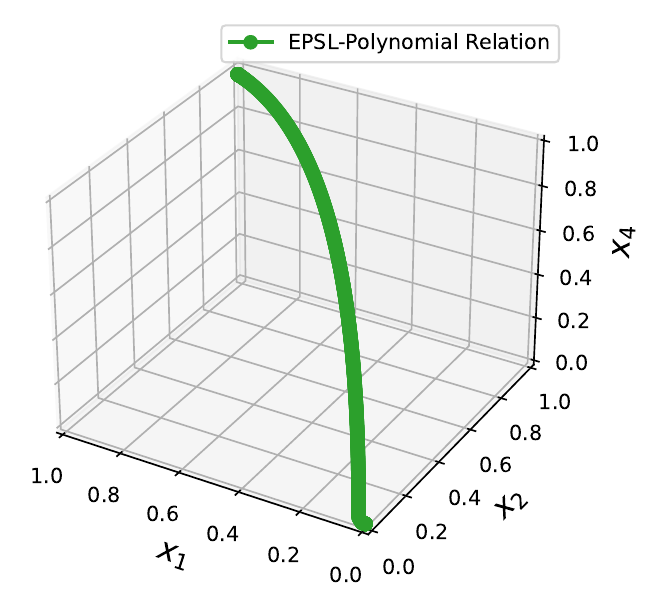}}
\hfill
\subfloat[PF by EPSL]{\includegraphics[width = 0.25\linewidth]{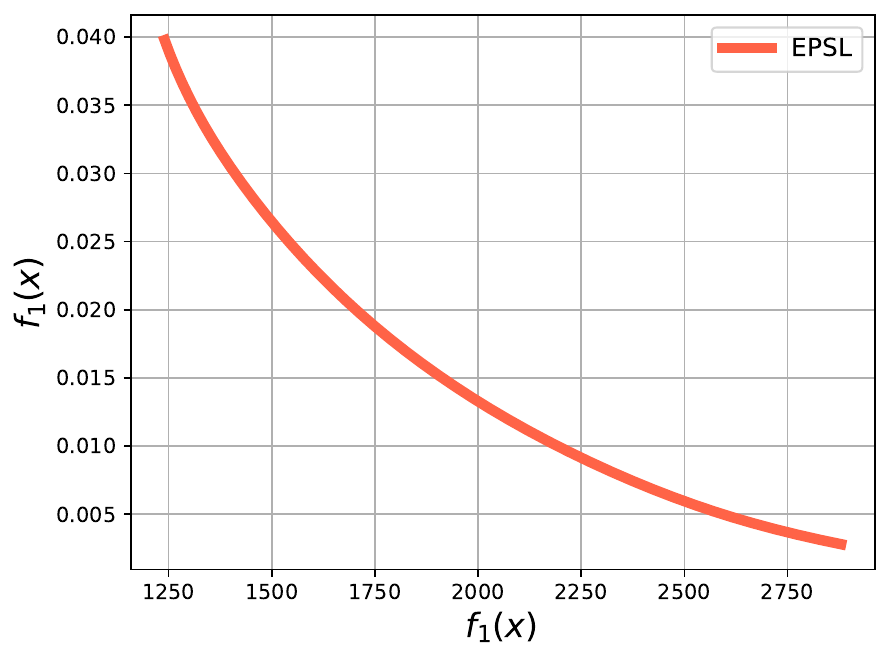}}
\hfill
\subfloat[PF by EPSL with Polynomial Shape]{\includegraphics[width = 0.25\linewidth]{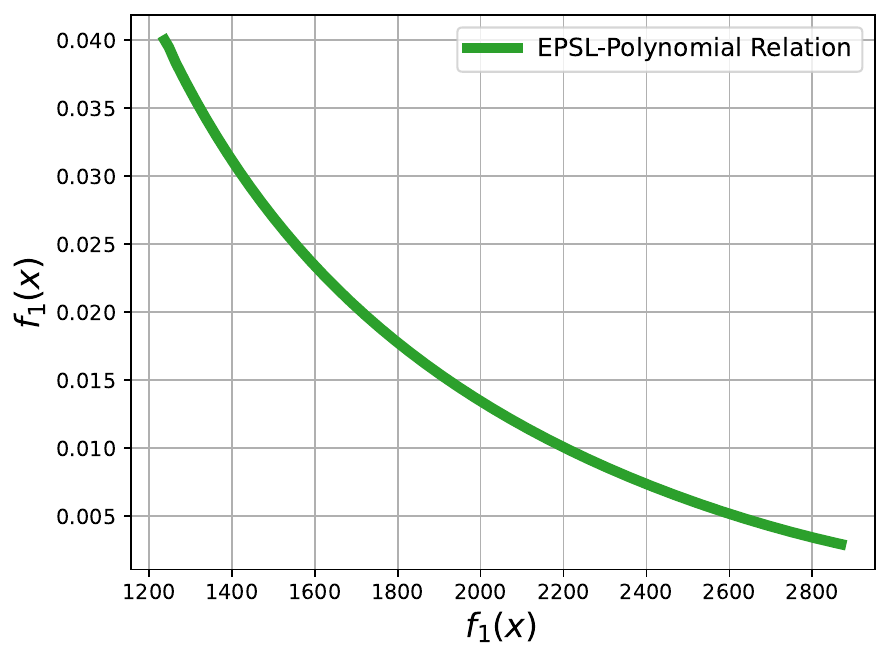}}
\caption{\textbf{EPSL with Learnable Polynomial Relationship for the Four Bar Truss Design Problem:} \textbf{(a)} The approximate Pareto set learned by the original EPSL. \textbf{(b)} The approximate Pareto set with learnable polynomial relation. \textbf{(c)} The approximate Pareto front obtained by the original EPSL. \textbf{(d)} The approximate Pareto front corresponds to the learned Pareto set with the polynomial structure.}
\label{fig_epsl_variable_relation_re21}
\end{figure*}

\subsection{Optimal Set with Shared Components}

To validate the ability of EPSL to learn the optimal solution set with shared components, we conduct experiments on three different engineering design problems. For the four bar truss design problem (RE21)~\cite{cheng1999generalized}, as shown in Fig.~\ref{fig_results_re21_shared_component}, we learn the optimal solution sets with a shared length of different bars ($\vx_1$, $\vx_2$, $\vx_3$, $\vx_4$) to minimize the structural volume ($f_1(\vx)$) and the joint
displacement ($f_2(\vx)$). For the reinforced concrete beam design problem (RE22)~\cite{amir1989nonlinear}, as shown in Fig.~\ref{fig_results_re23_shared_component}, we consider the optimal solution sets that share the area of the reinforcement ($\vx_1$), the width of the beam ($\vx_2$), the depth of the beam ($\vx_3$) or both the width and depth of the beam ($\vx_2$ and $\vx_3$), respectively. The goal of RE22 is to minimize the total cost of concrete and reinforcing steel ($f_1(\vx)$) and total constraint violation ($f_2(\vx)$). By sharing different components, each learned solution set has its own structure. For some applications, the solution sets with shared components might perform in a similar way to the unconstrained Pareto set, such as sharing the length of a specific bar for the four bar truss design (e.g., Fig.~\ref{fig_results_re21_shared_component}(d)). In other cases, the shared component might degenerate or sometimes significantly worsen the performance. These learned solution sets can provide valuable knowledge to support informative decision making.

For the rocket injector design problem (RE37)~\cite{vaidyanathan2003cfd}, the solution sets with shared hydrogen flow angle ($\vx_1$), hydrogen area ($\vx_2$), oxygen area ($\vx_3$), or oxidizer post tip thickness ($\vx_4$) are learned, respectively, as shown in Fig.~\ref{fig_results_re37_shared_component}. The optimization objectives are to minimize the
volume ($f_1(\vx)$) and the stress in a gear shaft ($f_2(\vx)$) of a speed reducer, as well as the total constraint violation ($f_3(\vx)$). According to the results, the solution sets with different shared components can approximately cover different subregions of the Pareto set. In addition, the learned solution set with shared oxidizer post tip thickness ($\vx_4$) is nearly identical to the learned Pareto set without any structure constraint. This means that we can have a simpler and regularized solution set (e.g., all solutions have the same value of $\vx_4$) without sacrificing the optimization performance. The results in this subsection clearly show that EPSL can successfully learn the solution sets with various shared component constraints, which allows decision-makers to trade off the Pareto optimality with their desirable solution set structures.

\subsection{EPSL with Learnable Variable Relationship}

In this subsection, we investigate EPSL's ability to learn an optimal solution set with a predefined and learnable variable relationship for the synthetic optimization problem and the real-world four-bar truss design problem (RE21)~\cite{cheng1999generalized}. For the synthetic optimization problem (\ref{synthetic_problem}), the ground truth optimal Pareto set satisfies the following formulation:
\begin{equation}
\vx_i = \sin(10(\vx_1 - 0.5)), \quad \forall i \in [2,\ldots, n]
\label{synthetic_problem_ps}
\end{equation}
which is unknown for the algorithm.  Decision-makers typically have a finite set of approximate solutions with traditional MOEAs (e.g., Fig.~\ref{fig_moea_epsl}(a)) or a somehow black-box model-based Pareto set approximation with the original EPSL (e.g., Fig.~\ref{fig_epsl_variable_relation_syn}(a)).

Suppose some prior knowledge of the Pareto set, such as the sine-curve-like structure, is available (e.g., via preliminary investigation), we can directly incorporate them into the Pareto set modeling:
\begin{align}
&\vx_1=h^{\vp}_{\vtheta}(\vlambda), \nonumber \\
&\vx_i =  \sin(\valpha_i(\vx_1 - \vbeta_i)), \forall i \in [2,\ldots, n],
\label{model_syn_sin}
\end{align}
where $h^{\vp}_{\vtheta}(\vlambda)$ is the Pareto set model with learnable parameter $\vtheta$ and $\vx_i =  \sin(\valpha_i(\vx_1 - \vbeta_i))$ is the predefined variable relationship with learnable parameters $\valpha$ and $\vbeta$. The parameters $[\vtheta, \valpha, \vbeta]$ can be easily optimized by our proposed evolutionary stochastic optimization method in \textbf{Algorithm~\ref{alg_epsl_framework}}. According to the results shown in Figure~\ref{fig_epsl_variable_relation_syn}, the learned Pareto set can match the ground truth Pareto set while providing more useful information (e.g., the explicit parametric formulation) for the given problem. This extra information could be useful to support more informative decision-making.

For the real-world four-bar truss design problem (RE21), the formulation of ground truth Pareto set is completely unknown. In this case, we propose a simple polynomial formulation to approximate the Pareto set:
\begin{align}
&\vx_1=h^{\vp}_{\vtheta}(\vlambda), \nonumber \\
&\vx_i =  1 - \valpha_i (\vx_1 - \vbeta_i)^2, \forall i \in [2,\ldots, n].
\label{model_re21_poly2}
\end{align}
By optimizing the learnable parameters $[\vtheta, \valpha, \vbeta]$, we obtain the parametric Pareto set approximation as in Figure~\ref{fig_epsl_variable_relation_re21}. While the learned simple parametric Pareto set is not identical to the unconstrained approximate Pareto set by the original EPSL, their corresponding Pareto fronts are very similar to each other. With a similar optimization performance, the much simpler parametric solution set could be preferred for practical decision-making.

\subsection{EPSL with Shape Structure Constraint}

\begin{figure}[t]
\centering
\subfloat[PS: Synthetic Problem]{\includegraphics[width = 0.5\linewidth]{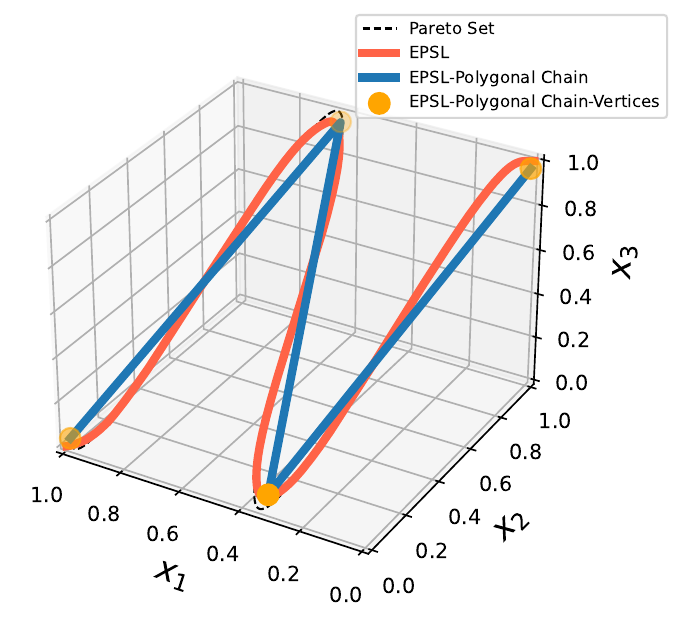}}
\subfloat[PF: Synthetic Problem]{\includegraphics[width = 0.5\linewidth]{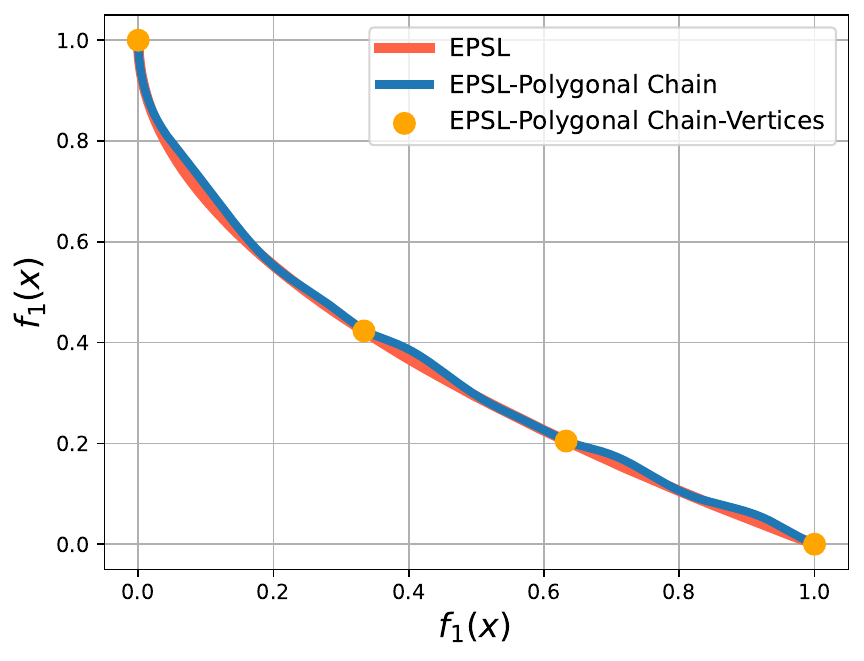}} \\
\subfloat[PS: Four Bar Truss Design]{\includegraphics[width = 0.5\linewidth]{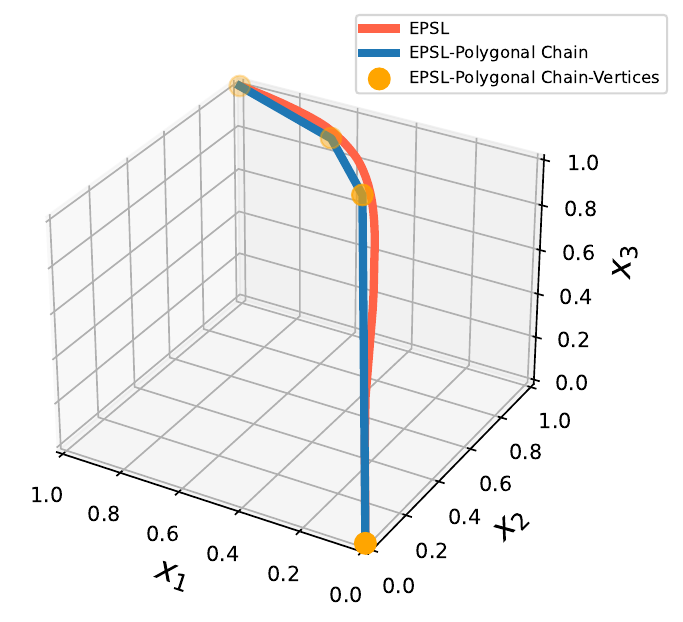}}
\subfloat[PF: Four Bar Truss Design]{\includegraphics[width = 0.5\linewidth]{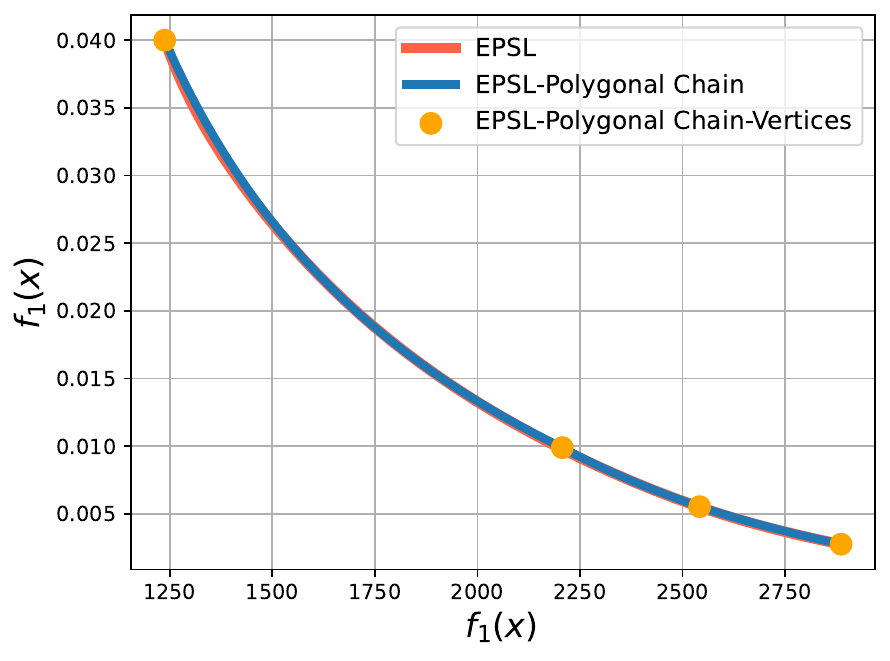}}
\caption{\textbf{EPSL with Polygonal Chain Structure Constraints:}  \textbf{(a)(c)} EPSL can successfully learn the polygonal chains with crucial vertices to well approximate the ground truth Pareto sets. \textbf{(b)(d)} Even though the constrained Pareto sets have much simpler structures, their corresponding Pareto fronts can provide nearly comparable trade-offs with the ground truth Pareto fronts.}
\label{fig_syn_sin_key_points}
\end{figure}

This subsection demonstrates the ability of EPSL to learn a Pareto set with a simple polygonal chain structure. We set the number of vertices $K = 4$ for all problems, and compare its performance with the unconstrained EPSL. The learned Pareto sets and the corresponding Pareto front on the synthetic problem (\ref{synthetic_problem}) and the four bar truss design problem (RE21) are shown in Fig.~\ref{fig_syn_sin_key_points}. According to the results, EPSL with polygonal chain constraints can find a solution set with a much simpler structure, while its corresponding Pareto front performs slightly worse.

We also report the hypervolume difference ($\Delta$ HV) for other bi-objective engineering design problems in Table~\ref{table_epml_polygonal_chain}. These results show that EPSL can find a reasonably good approximation to the ground truth Pareto set with a much simpler polygonal chain structure. A simple Pareto set with an explainable structure, instead of a complicated one, could be easier to understand and can shed insights into the optimization problem at hand. In some applications, decision-makers could prefer such a simple Pareto set while it might be trade-offed with the absolute Pareto optimality. Our proposed EPSL can well support this demand.

\begin{table}[t]
\centering
\caption{The Median Hypervolume GAP ($\Delta$ HV) of EPSL with and without polygonal chain structure. Significantly better results are highlighted in bold.}
\label{table_epml_polygonal_chain}
\begin{tabular}{lcc}
\hline
\multicolumn{1}{c}{} & EPSL-Polygonal Chain & EPSL              \\ \hline
Syn                  & 8.42e-03             & \textbf{3.67e-03} \\
RE21                 & 8.23e-03             & \textbf{3.91e-04} \\
RE22                 & 7.06e-02             & \textbf{6.47e-03} \\
RE23                 & 5.08e-02             & \textbf{1.42e-03} \\
RE24                 & 1.93e-01             & \textbf{7.07e-05} \\
RE25                 & 4.22e-01             & \textbf{6.98e-08} \\ \hline
\end{tabular}
\end{table}

\section{Discussion and Conclusion}
\label{sec_conclusion}

This paper has proposed a simple yet efficient evolutionary Pareto set learning (EPSL) method to incorporate structure constraints on the whole solution set for multiobjective optimization. The proposed method generalizes the two major components of MOEA/D, namely decomposition and collaboration, with the learning-based model. For decomposition, EPSL generalizes the scalarization approach from a fixed and finite set of preferences to consider all infinite valid preferences via a single neural network model. For collaboration, we propose an efficient evolutionary stochastic gradient search method to optimize the learned Pareto set concerning all preferences and structure constraints simultaneously. The knowledge and information are now shared among all preferences within the same model, rather than exchanged among a fixed finite set of subproblems.  It allows decision-makers to trade off the optimal objective values with desirable properties on the solution set, which goes beyond the classic Pareto optimality for multiobjective optimization.

In terms of computational overhead, our proposed EPSL is about the same as a single run of MOEA/D, while it allows decision makers to choose any trade-off solutions on the whole learned Pareto set. We believe it is a promising alternative to the traditional MOEAs and can lead to many interesting applications and algorithms. In this paper, we focus on the key idea of learning the whole Pareto set with structure constraints via a single model, and keep the model and optimization algorithm simple and straightforward. Many important improvements should be investigated in future work:
\begin{itemize}

    \item Building more powerful models to approximate complicated Pareto sets and handle complex structure constraints, such as those with discrete decision variables, disconnected structures, or complex patterns.
    
    \item Proposing more efficient optimization methods for learning the Pareto set model, which should go beyond the straightforward Monte-Carlo sampling and simple evolutionary gradient descent.
    
    \item Generalizing EPSL with structure constraints to tackle different optimization scenarios such as expensive optimization, interactive optimization, and dynamic optimization.
    
\end{itemize}

\bibliographystyle{Style/IEEEtran}
\bibliography{ref_multiobjective_optimization,ref_bayesian_optimization, ref_multi_task_learning, ref_combinatorial_optimization,
ref_learning_to_optimize}
\end{document}